\documentclass[a4paper,fleqn]{cas-dc}
\usepackage[authoryear]{natbib}

\usepackage{float}
\usepackage{multirow}

\usepackage{subcaption}
\usepackage{multirow}
\usepackage{adjustbox}
\usepackage{amsmath,amsfonts,bm}
\usepackage{alphalph}
\usepackage{amssymb}
\usepackage{pifont}
\usepackage{comment}
\usepackage{amssymb} 
\usepackage{float}
\usepackage{algorithm}
\usepackage{algpseudocode}
\usepackage{xcolor}
\usepackage{pifont}
\usepackage{tabularx}
\usepackage{booktabs}

\def\tsc#1{\csdef{#1}{\textsc{\lowercase{#1}}\xspace}}
\tsc{WGM}
\tsc{QE}
\tsc{EP}
\tsc{PMS}
\tsc{BEC}
\tsc{DE}

\def\eqref#1{equation~\ref{#1}}

\def\1{\bm{1}}

\newcommand{\model}{Smart Transfer}

\begin{document}
\let\WriteBookmarks\relax
\def\floatpagepagefraction{1}
\def\textpagefraction{.001}

\shorttitle{\model}

\shortauthors{Li et~al.}

\title [mode = title]{Smart Transfer: Leveraging Vision Foundation Model for Rapid Building Damage Mapping with Post-Earthquake VHR Imagery}                      



\begin{abstract}
Living in a changing climate, human society now faces more frequent and severe natural disasters than ever before. As a consequence, rapid disaster response during the ``Golden 72 Hours'' of search and rescue becomes a vital humanitarian necessity and community concern. However, traditional disaster damage surveys routinely fail to generalize across distinct urban morphologies and new disaster events. Effective damage mapping typically requires exhaustive and time-consuming manual data annotation. To address this issue, we introduce Smart Transfer, a novel Geospatial Artificial Intelligence (GeoAI) framework, leveraging state-of-the-art vision Foundation Models (FMs) for rapid building damage mapping with post-earthquake Very High Resolution (VHR) imagery. Specifically, we design two novel model transfer strategies: first, Pixel-wise Clustering (PC),  ensuring robust prototype-level global feature alignment; second, a Distance-Penalized Triplet (DPT), integrating patch-level spatial autocorrelation patterns by assigning stronger penalties to semantically inconsistent yet spatially adjacent patches. Extensive experiments and ablations from the recent 2023 Türkiye-Syria earthquake show promising performance in multiple cross-region transfer settings, namely Leave One Domain Out (LODO) and Specific Source Domain Combination (SSDC). Moreover, Smart Transfer provides a scalable, automated GeoAI solution to accelerate building damage mapping and support rapid disaster response, offering new opportunities to enhance disaster resilience in climate-vulnerable regions and communities. The data and code are publicly available at \url{https://github.com/ai4city-hkust/SmartTransfer}.
\end{abstract}



\author[1]{Hao Li}[type=editor,
                        auid=000,
                        bioid=1,
                        orcid=0000-0002-6336-8772]

\cormark[1]
\ead{hao.li@nus.edu.sg}

\author[2]{Liwei Zou}
\ead{lzou260@connect.hkust-gz.edu.cn}

\author[3]{Wenping Yin}[type=editor,
                        auid=000,
                        bioid=1,
                        ]
\ead{yin@cumt.edu.cn}

\author[4]{Gulsen Taskin}
\ead{gulsen.taskin@itu.edu.tr}

\author[5]{Naoto Yokoya}
\ead{yokoya@k.u-tokyo.ac.jp}

\author[6]{Danfeng Hong}
\ead{hongdanfeng1989@gmail.com}

\author[2]{Wufan Zhao}

\cormark[1]
\ead{wufanzhao@hkust-gz.edu.cn}

\affiliation[1]{organization={Department of Geography, National University of Singapore},
    city={Singapore},
    postcode={117570}, 
    country={Singapore}
    }

\affiliation[2]{organization={Urban Governance and Design Thrust, The Hong Kong University of Science and Technology (Guangzhou)},
    city={Guangzhou},
    postcode={511453}, 
    state={Guangdong},
    country={China}
    }

\affiliation[3]{organization={School of Environment and Spatial Informatics, China University of Mining and Technology},
    city={Xuzhou},
    postcode={221116}, 
    state={Jiangsu},
    country={China}}

\affiliation[4]{organization={Disaster Management Institute, Istanbul Technical University},
    city={Istanbul},
    postcode={34469}, 
    state={Sariyer},
    country={Türkiye}}

\affiliation[5]{organization={Department of Complexity Science and Engineering, University of Tokyo},
    city={Chiba},
    postcode={277-8561}, 
    country={Japan}}

\affiliation[6]{organization={School of Automation, Southeast University},
    city={Nanjing},
    postcode={211189}, 
    state={Jiangsu},
    country={China}}

\cortext[cor1]{Corresponding authors: Wufan Zhao and Hao Li}


\maketitle

\section{Introduction}

In the recent decade, the frequency and intensity of natural disasters, such as floods \citep{li2020exploration, wagner2026fully}, earthquakes \citep{anniballe2018earthquake, yunus2020decadal}, hurricanes \citep{zheng2025nighttime, li2025cross}, heatwaves \citep{manoli2019magnitude, zhao2023understanding}, and wildfires \citep{zhang2021learning, marcos2023assessing} have increased significantly, driven by climate change and expanding human activities. The Emergency Events Database recorded $399$ natural hazard-related disasters in $2023$ alone, resulting in $86,473$ fatalities and approximately $202.7$ billion dollars in economic losses, with the Türkiye-Syria earthquake among the most devastating events \citep{delforge2025dat, zhao2021building}. This trend has increased pressure on global communities, where limited resources struggle to keep pace with growing damages, contributing to the ``Disaster Resilience Gap'', which is further exacerbated by the digital divide between the Global North and South \citep{ghosh2023very,hallegatte2019lifelines}.


Earth Observation (EO) data, featured by the wide spatial coverage and increasing spatiotemporal resolution, have become a critical data source for large-scale disaster monitoring and rapid post-event assessment \citep{chen2019change, bai2023deep,tacskin2021review}. Modern EO constellations, encompassing aerial, space, and ground systems, have substantially increased the temporal and spatial resolution of disaster monitoring, reducing response times to hours or near real-time. To this end, EO plays an increasingly important role in shaping disaster risk reduction strategies. For instance, the Sendai Framework outlines action priorities including “understanding disaster risks”, “investing in disaster reduction for resilience”, and “enhancing disaster preparedness”, which benefit from enhanced EO capabilities \citep{le2020space}. However, the increasing volume and velocity of EO data generated during and after disaster events far exceed the processing capacity of classic EO analytic pipelines, putting a pressing need for more automated and scalable disaster mapping solutions. 

\begin{figure}
\centering
\centering
\includegraphics[width=\linewidth]{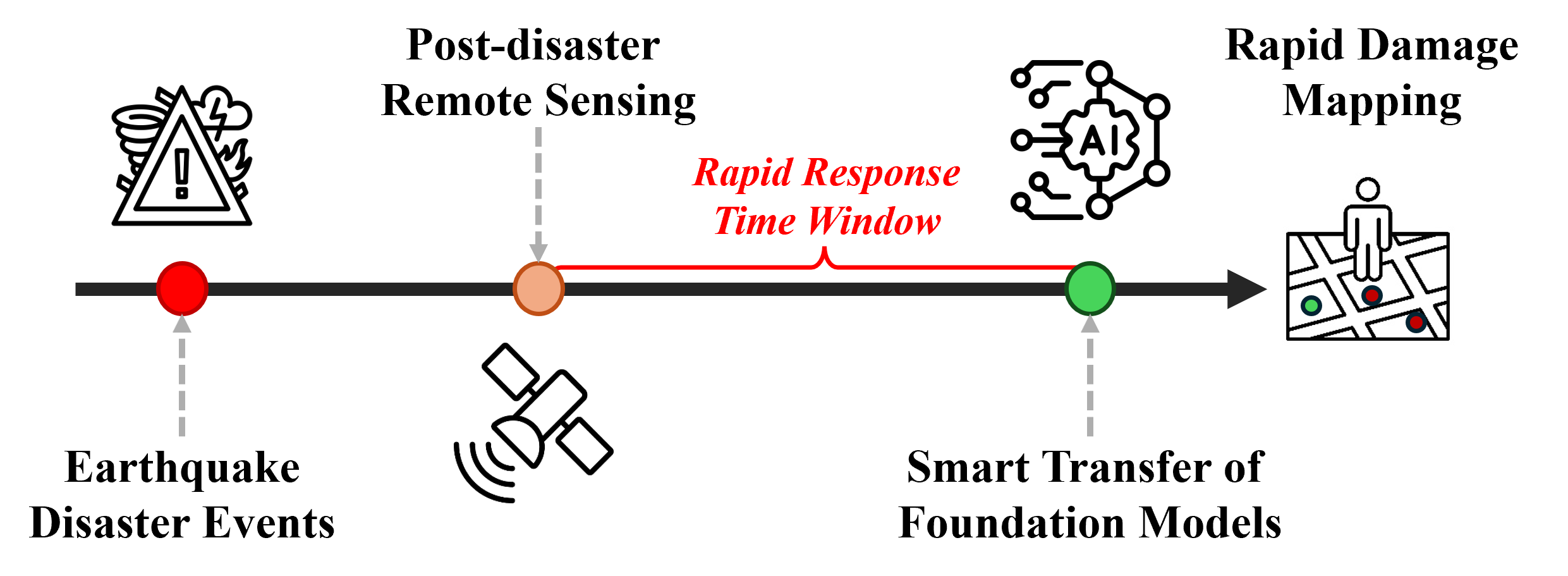}
\caption{A conceptual diagram of post-earthquake damage mapping, illustrating the Smart Transfer framework's objective to minimize the rapid response time window for critical disaster response operations.}
\label{fig:diagram}
\end{figure}

The paradigm of disaster mapping is undergoing a significant technology shift from classic field-based surveys to automated Geospatial Artificial Intelligence (GeoAI) approaches \citep{hu2019geoai,liu2024explainable, li2024geoai}. Although traditional field-based damage assessment remains fundamental for detailed structural reconnaissance, it is inherently constrained by site accessibility, human perception, and the massive time requirements that hinder the ``Golden 72 Hours'' of search and rescue operations in disaster response, as shown in Figure \ref{fig:diagram}. Recent works demonstrate the effectiveness and adaptability of GeoAI approaches in mapping disaster damage using EO data within hours of events (e.g., flooding, wildfires, hurricanes, earthquakes, etc.) \citep{yokoya2020breaking, gong2025urban, li2025cross, yin2025triple, li2026buildingmultiview}. Despite these advances, existing GeoAI methods largely rely on massive amounts of labeled data and region-specific training, which limits their scalability and rapid deployment in new disaster scenarios. 

The emergence of vision Foundation Models (FMs), such as the Segment Anything Model (SAM) and geospatial-specific models like \textit{Prithvi}, represents the latest transformative advancement in AI for EO \citep{awais2025foundation, kirillov2023segment, zhao2022extracting, hong2024multimodal,hong2026foundation}. Unlike traditional Convolutional Neural Networks (CNNs) trained for specific tasks, FMs learn robust, general-purpose representations of spatial and spectral features \citep{mai2024opportunities, li2026adapting, hong2026hyperspectral} that can be adapted to diverse downstream tasks with limited training data, such as 3D city reconstruction \citep{hua2025sat2city}, urban built environment assessment \citep{lu2024transferable}, and spatial-temporal mobility prediction \citep{xu2025predicting}. This capability is particularly relevant for disaster mapping, where rapid deployment is required but labeled data and human resources are often limited, and damage characteristics vary significantly across regions and events \citep{li2023rethink, hong2023cross, zheng2024towards}. 

Despite the availability of large-scale benchmark datasets such as \textit{xBD} \citep{gupta2019xbd} and \textit{BRIGHT} \citep{chen2025bright}, directly applying vision FMs to disaster damage assessment remains challenging, as existing models are not specifically tailored to various damage patterns and labeled data remain scarce in real-world scenarios. This highlights the pressing need for smart transfer approaches that can effectively leverage the representation capability of FMs while incorporating domain knowledge to support cross-regional disaster mapping under limited supervision. Fortunately, the excellent generalizability of the vision FMs accumulated a nice momentum towards cross-regional building damage mapping, which just needs to ``warm up'' in the targeted disaster scenes guided by geographical and task-specific knowledge (e.g., building damage classification) before reaching a highly competitive performance level compared to its counterpart of non-FMs trained for the specific task.

In this paper, we propose a novel foundation model-based  GeoAI framework, namely \textit{Smart Transfer}, leveraging prototype clustering and spatial autocorrelation to adapt state-of-the-art vision FM backbones for robust cross-regional building damage mapping with minimal training and ground truth labels. To validate the Smart Transfer framework, the 2023 Türkiye-Syria earthquake, consisting of nine distinct regions, is selected as a case study. The complex urban environment of southeastern Türkiye, coupled with common building-level soft-story vulnerabilities and diverse urban morphological characteristics, makes this dataset a rigorous real-world benchmark for validating the proposed framework. By combining Very High Resolution (VHR) satellite imagery and open-access building footprints, the Smart Transfer framework designs and tests two novel FM transfer strategies: Pixel-wise Clustering (PC) and Distance-Penalized Triplet (DPT) to handle this substantial domain shift across diverse urban morphologies and damage patterns. To the best of our knowledge, Smart Transfer is one of the first works to explore the generalizability of vision FMs for disaster mapping, providing timely insights into how to close the disaster resilience gap and shorten the rapid response time window. The scientific contributions of Smart Transfer are summarized as follows:

\begin{itemize}
    \item To design two novel Smart Transfer strategies, namely \textit{PC} and \textit{DPT}, to leverage vision FMs for cross-region building damage assessment with minimal effort in additional training and labeling.
    \item To benchmark the Smart Transfer framework under distinct cross-region transfer settings, specifically Leave One Domain Out (LODO) and Specific Source Domain Combination (SSDC), in facilitating various rapid disaster response scenarios.
    \item To contribute a high-quality and large-scale building damage assessment dataset, including post-disaster VHR imagery and open access building footprints, together with open source code for vision FMs to the disaster mapping community.
\end{itemize}

The remainder of this paper is organized as follows: Section 2 describes the study area and the geospatial datasets used in this work. Section 3 describes the proposed damage mapping framework together with the zero-shot and few-shot learning approaches. Section 4 presents experimental results, including an extensive ablation study, to validate the effectiveness and robustness of Smart Transfer. Section 5 discusses key findings as well as limitations and future directions, and Section 6 concludes the paper by highlighting the main contributions.

\section{Study area and data}\label{studyarea}

\begin{figure*}[!ht]
\centering
\centering
\includegraphics[width=\linewidth]{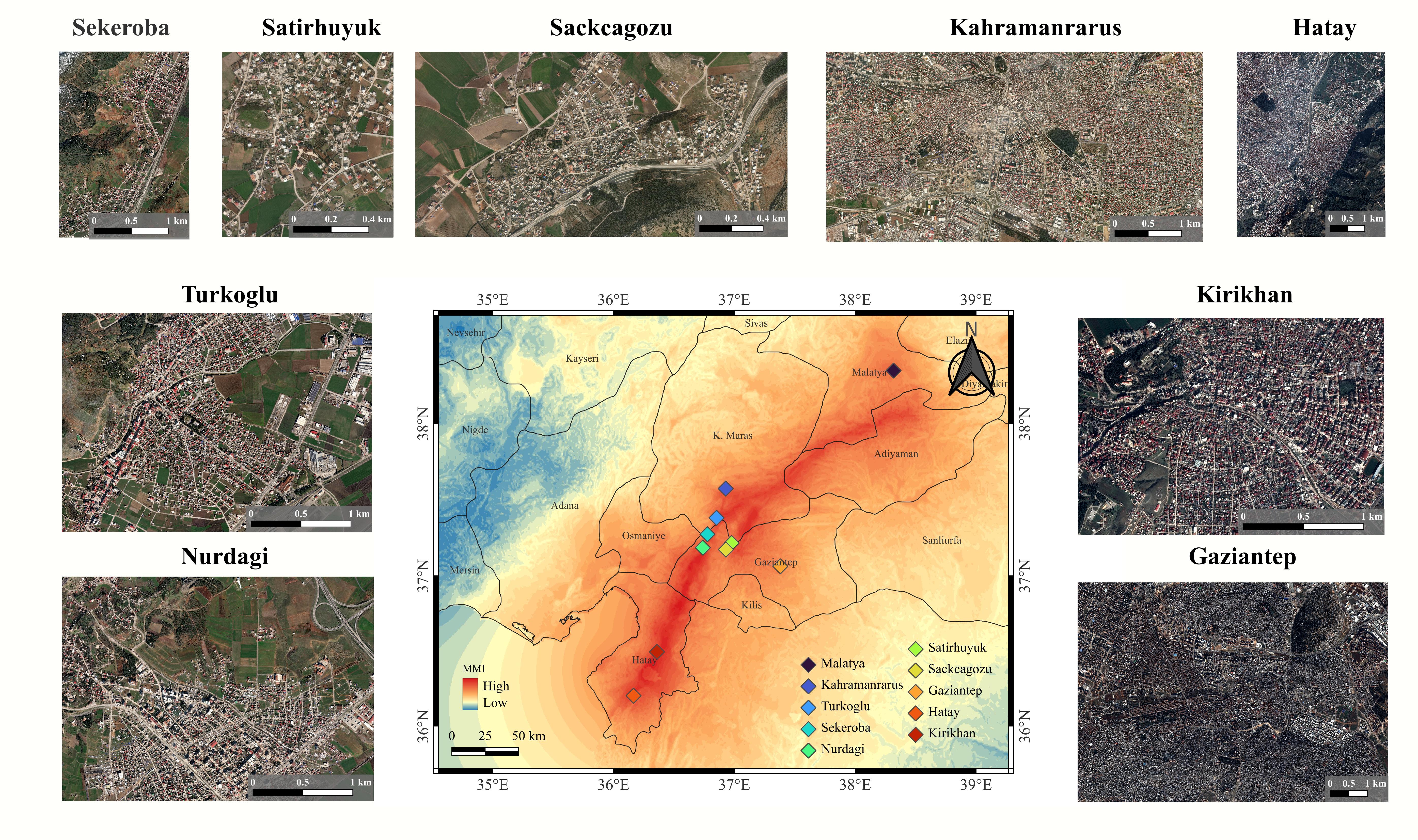}
\caption{Study area overview displaying the USGS Modified Mercalli Intensity (MMI) shakemap. Post-disaster VHR imagery is provided for the 9 selected urban regions.}
\label{fig:study_area}
\end{figure*}

\begin{table*}[!ht]
\centering
\caption{Dataset statistics across the 9 selected study regions, including the distribution of total VHR image tiles, building footprints, and ground truth annotations.}
\label{tab:dataset_stats}
\begin{tabular}{lrrrrr}
\toprule
Region & Area (km$^2$) & Total Tiles & Total Buildings & Labeled Tiles & Labeled Buildings \\
\midrule
Gaziantep     & 123.03 & 30,155 & 173,774 & 42  & 45  \\
Hatay         & 105.30 & 25,781 & 47,752  & 210 & 258 \\
Kahramanmaras & 27.41  & 18,020 & 40,419  & 414 & 419 \\
Kirikhan      & 12.18  & 3,016   & 11,958  & 53  & 51  \\
Nurdagi       & 15.21  & 10,032   & 4,076   & 225 & 176 \\
Sakcagozu    & 4.50   & 3,024   & 1,196   & 74  & 79  \\
Satirhuyuk    & 4.11   & 2,745   & 907     & 148 & 131 \\
Sekeroba      & 22.52  & 14,841   & 4,431   & 97  & 79  \\
Turkoglu      & 17.13  & 11,400   & 4,054   & 77  & 45  \\
\midrule
All Regions   & 331.37 & 119,014 & 288,567 & 1,340 & 1,283 \\
\bottomrule
\end{tabular}
\end{table*}

\subsection{Study area}

The study area covers the region most severely affected by the 7.8 Mw earthquake that struck southeastern Turkey on 6 February 2023. The earthquake ruptured a major portion of the East Anatolian Fault, producing strong ground shaking across a broad swath of the provinces of Kahramanmaras, Hatay, Osmaniye, Adıyaman, Gaziantep, and Malatya. As shown in Figure \ref{fig:study_area}, the USGS shakemap indicates a continuous belt of high Modified Mercalli Intensity (MMI) values along the fault rupture, reflecting severe to destructive shaking levels.

\begin{figure*}[t!]
\centering
\centering
\includegraphics[width=\linewidth]{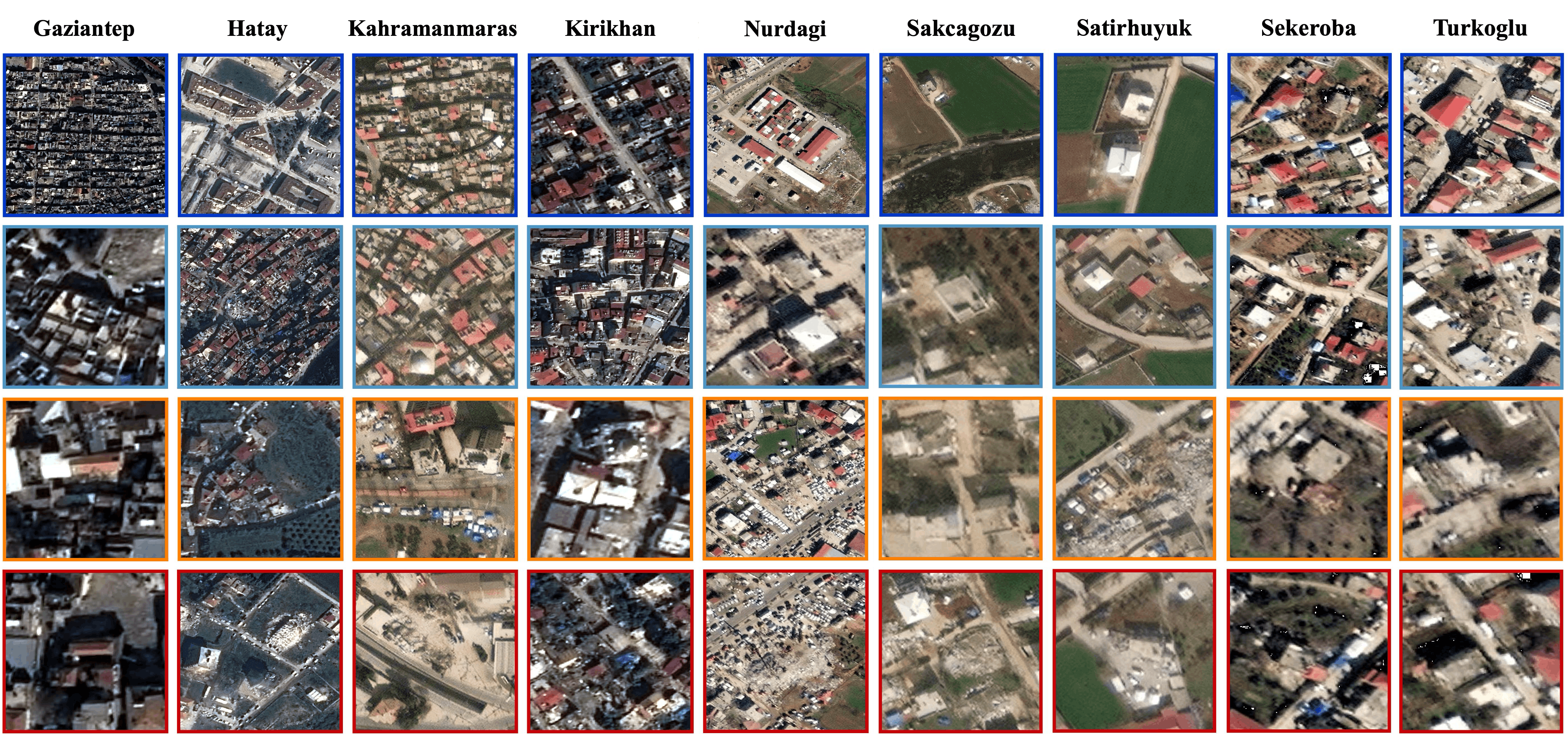}
\caption{Selected examples of multiple building damage levels from 9 study regions, from top to bottom row refer to slight damage, heavy damage, buildings requiring demolition, and collapse building, respectively.}
\label{fig:damage_example}
\end{figure*}

Within this corridor of intense ground motion, we delineated 9 representative urban regions as shown in Figure~\ref{fig:study_area}, including Sekeroba, Satirhuyuk, Sakcagozu, Nurdagi, Kahramanmaras, Gaziantep, Hatay, Kirikhan, and Turkoglu. These locations spread from the near-fault, high-MMI belt to surrounding zones where shaking intensity gradually decreases, thereby capturing a continuous gradient from the most heavily impacted areas to moderately affected peripheral regions. This diverse set of study locations reflects the full range of ground-shaking conditions during the 2023 Turkey earthquake sequence and provides a solid foundation for evaluating earthquake-induced urban damage through rapid, high-resolution mapping supported by the proposed Smart Transfer framework.

\subsection{Data source}

To support reproducibility and future research, we made all data and code publicly available, including post-earthquake VHR imagery \footnote{\url{https://github.com/ai4city-hkust/SmartTransfer}}. In the rest of this section, we elaborate on the details of data processing. 

\subsubsection{Post-disaster VHR remote sensing data}
The post-disaster VHR imagery used in this study was acquired by the\textit{ Pléiades 1A} and \textit{1B} satellite constellation, operated by Airbus Defence and Space. The dataset provides optical imagery with a ground sampling distance of $0.3 $m, enabling detailed characterization of individual buildings, road networks, and debris distributions. As one of the few sub-meter satellite datasets captured shortly after the 2023 Türkiye–Syria earthquake, the Pléiades imagery offers exceptionally rich spatial detail for damage interpretation and semantic segmentation. Its coverage spans a diverse range of affected environments, from dense urban centers to sparsely populated rural settlements, making it a critical data source for high-precision post-disaster mapping and analysis.

\begin{figure*}[t!]
\centering
\centering
\includegraphics[width=0.8\linewidth]{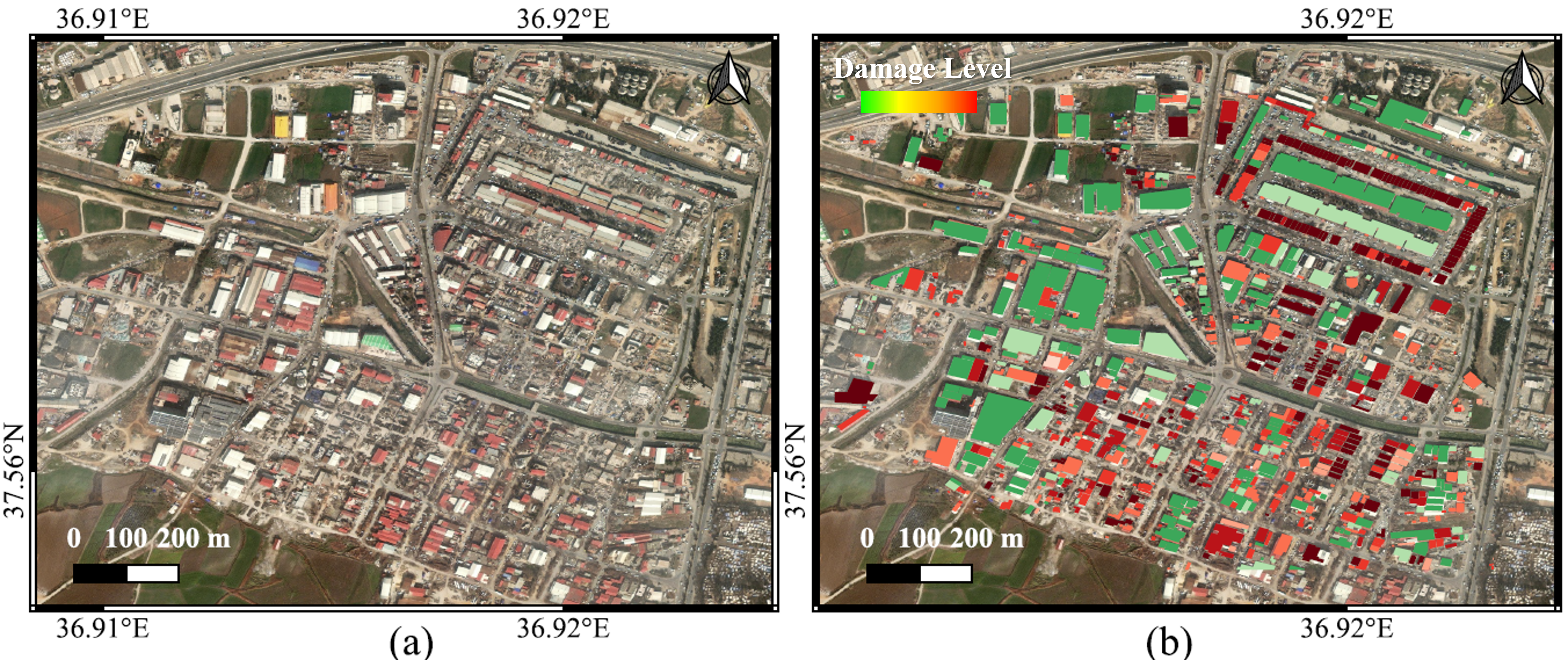}
\caption{Post-disaster remote sensing imagery of part of Kahramanmaras (a) post-disaster VHR imagery and (b) with annotated damaged buildings (red) and undamaged buildings (green). Color intensity indicates the damage score of the ground truth annotation.}
\label{fig:kah_annotated}
\end{figure*}

\subsubsection{Pre-disaster building footprint data}
The pre-disaster building footprint data is open-access from the GlobalBuildingAtlas \citep{zhu2025globalbuildingatlas}, providing vectorized representations of individual building outlines. These building polygons were compiled from existing national and global-scale geospatial products that offer reliable coverage of urban and peri-urban areas. The dataset delineates the horizontal extent of structures at the building-instance level, supplying essential baseline information for identifying structural losses when compared with post-disaster imagery. The footprints serve as a stable pre-disaster reference for building damage classification and segmentation, ensuring that building-level analyses are grounded in consistent spatial geometry independent of debris, occlusion, or collapse patterns visible in the post-disaster VHR imagery.

\subsubsection{Damaged-building annotation benchmark}
The building damage annotation data used in this study were sourced from the publicly released post-disaster  \textit{KATE-CD dataset }curated by \cite{musaouglu2025kate} and further extended by the authors. A detailed statistic for each region is shown in Table \ref{tab:dataset_stats}. To ensure the rigor of the assessment, this annotation dataset is derived from VHR imagery acquired over earthquake-affected regions in Türkiye. The raw satellite images were divided into fixed-size tiles and manually annotated using the \textit{Label Studio} platform. The annotations focus on visible post-earthquake damage indicators, including roof collapses, structural deformations, and debris accumulation, and damaged buildings are precisely delineated using polygon geometries. As a result, we organized an expert mapping campaign where all $1,340$ labeled tiles across 9 regions were checked by at least two building-damage assessment experts. As shown in Figure \ref{fig:kah_annotated}, we conducted a wall-to-wall labeling for a subarea in Kahramanmaras, consisting of $1,020$ tiles and 900 building footprints to serve as a real-world deployment case. All labels were interpreted strictly from the imagery itself without reliance on auxiliary sources, ensuring the independence and objectivity of the annotations. As an openly accessible and fully reproducible resource, this dataset provides a reliable reference for training and evaluating building-damage identification methods in post-earthquake scenarios.

Figure \ref{fig:damage_example} provides representative annotation examples corresponding to different damage levels according to the crowdsourced MapRoulette projects, ranging from slightly damaged buildings to structures requiring demolition. These examples reveal both the visual diversity and the variability in annotation quality inherent in crowdsourced VGI, highlighting the limitations of relying solely on such data and motivating the need for more robust damage assessment methods.


\section{Methodology} \label{method}

\begin{figure*}[!t]
\centering
\centering
\includegraphics[width=\linewidth]{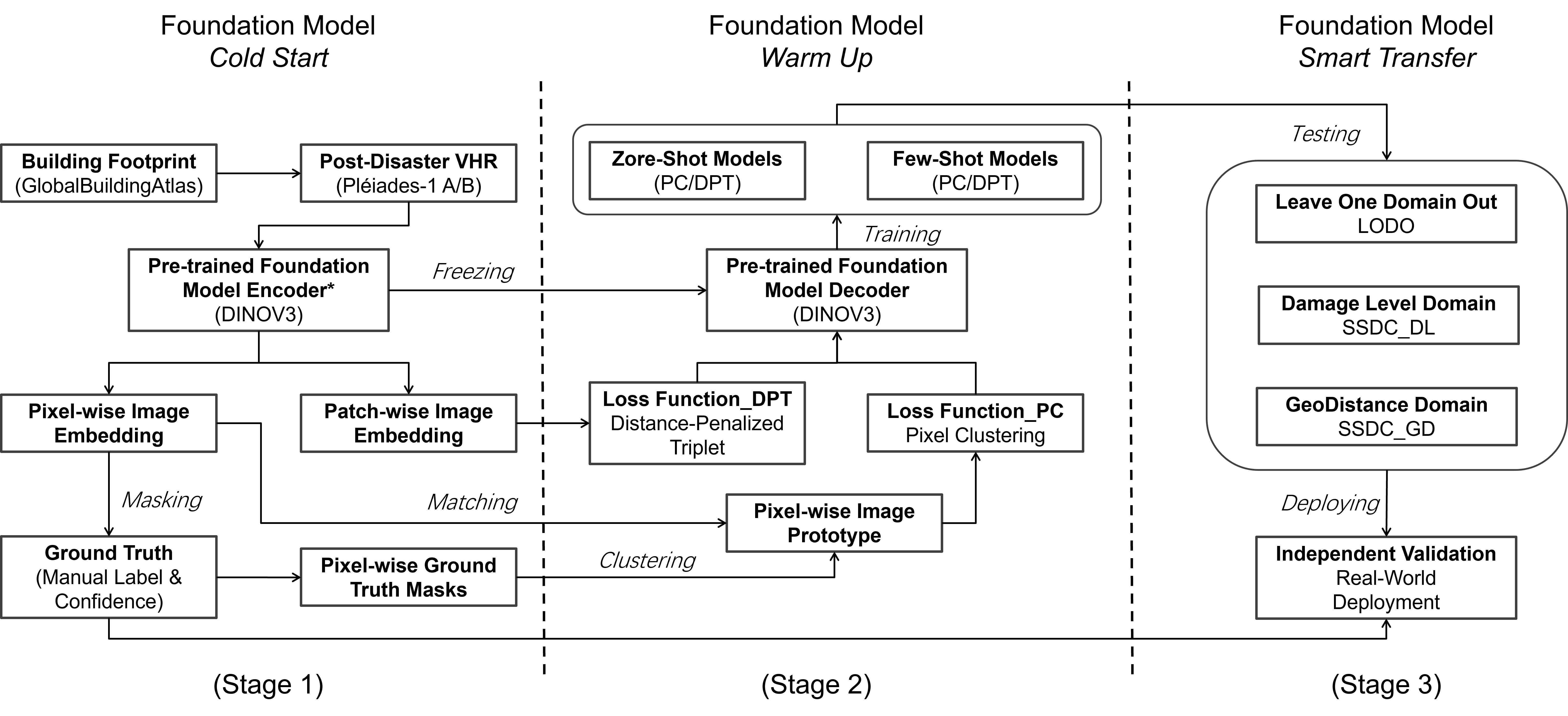}
\caption{Three Stages of Smart Transfer. Stage 1 is the cold start of foundation models, Stage 2 is the warm-up of foundation models, and Stage 3 is the smart transfer strategy designed for diverse damage mapping scenarios. }
\label{fig:flowchart}
\end{figure*}

\subsection{Backbone Foundation Model}

The overall pipeline of the proposed Smart Transfer framework is shown in Figure \ref{fig:flowchart}. We employ the ViT-Large (ViT-L/16) variant of DINOv3 pretrained on the large-scale SAT493M remote sensing dataset to balance computational capacity with feature richness. Given a post-event satellite image tile $x \in \mathbb{R}^{H \times W \times C}$ (where $H, W = 128$ and $C=3$ for RGB optical imagery), the image is tessellated into a grid of non-overlapping patches of size $P \times P$ (where $P=16$).
The number of patches (tokens) $N$ is given by:
$$N = \frac{H \cdot W}{P^2}.$$
Each 2D patch $x_p^{(i)}$ is flattened into a vector of dimension $P^2 \cdot C$ and linearly projected into the latent embedding dimension $D$ (where $D=1024$ for ViT-L) via a learnable projection matrix $\mathbf{E} \in \mathbb{R}^{(P^2 C) \times D}$.

\textbf{Pre-trained Foundation Model Vision Encoder:} The profound capability of DINOv3 to serve as a backbone FM vision encoder for rapid damage mapping lies in its self-supervised pre-training scheme \citep{simeoni2025dinov3,hong2024spectralgpt}. In this work, the smart transfer approach relies on the general-purpose vision features learned via a composite loss function $\mathcal{L}_{DINOv3}$ that combines global discrimination, local reconstruction, and dense feature preservation.

The pre-training of DINOv3 utilizes a multi-component objective function designed to balance global semantic understanding with dense local feature extraction \citep{simeoni2025dinov3}. The pre-training loss function is defined as:
\begin{equation}
    \mathcal{L}_{DINOv3} =  \mathcal{L}_{DINO} + \mathcal{L}_{iBOT} + \mathcal{L}_{Gram}.
\end{equation}

\textbf{Self-Distillation with DINO Loss:} First, the core objective is self-distillation using a Student-Teacher framework \citep{caron2021emerging}. The teacher network $g_{\theta_t}$ is an Exponential Moving Average of the Student $g_{\theta_s}$, providing stable learning targets. The model processes various ``masked views'' of post-disaster VHR imagery, noted as $V$, and the student learns to match the teacher's probability distribution $P$ across these views:
\begin{equation}
    \mathcal{L}_{DINO} = \sum_{x_g \in V} \sum_{x' \in V} H(P_t(x_g), P_s(x')),
\end{equation}
where $H$ denotes the cross-entropy. This self-distillation loss forces the model to learn high-level semantic representations that are invariant to scaling, cropping, and color differences. 

In rapid damage mapping, the DINO loss ensures that FM vision encoders can learn general-purpose features that are helpful for detecting building damages despite diverse damage characteristics, urban morphological characteristics, and geographical locations, serving as a solid foundation for the smart transfer approach.

\textbf{Masked Image Modeling with iBOT Loss:} Though the DINO loss optimizes the global image embeddings, the image BERT pre-training with Online Tokenizer (iBOT) component focuses on patch-level embeddings \citep{zhou2021ibot}. This is formulated as a Masked Image Modeling (MIM) task via distillation, where a portion of input patches is masked, and the student must predict the teacher's feature distribution for those specific obscured locations. Thus, the iBOT loss is defined as below:
\begin{equation}
    \mathcal{L}_{iBOT} = \sum_{x_g} \sum_{i \in Mask} H(P_t(x_g^{i}), P_s(x_s^{i})).
\end{equation}
The pre-training on this MIM task enables the FM vision encoder to learn local and spatial dependencies, for example, inferring a damaged roof patch from the neighborhood damaged wall tokens, which is critical for dense mapping tasks like damaged building segmentation in this work.

\textbf{Geometric Consistency with Gram Loss:} Gram anchoring loss further mitigates feature collapse effects where patch tokens become redundant and converge toward the global token \citep{cicchetti2024gramian}. This loss enforces a geometric constraint by aligning the pairwise correlation structure of the student network with that of a high-quality Gram Teacher. The Gram loss is defined by the squared Frobenius norm as below:
\begin{equation}
    \mathcal{L}_{Gram} = || G_s - G_g ||_F^2 = \sum_{i=1}^N \sum_{j=1}^N (G^{ij}_s - G^{ij}_g)^2,
\end{equation}
where $G_s, G_g \in \mathbb{R}^{N \times N}$ represent the Gram matrices for the student and teacher, respectively, while \textbf{$N$} is the number of patches.

By including the Gram loss, the FM vision encoder preserves the spatial and textural signatures of the urban fabric. For earthquake damage assessment, this ensures the encoder distinguishes between the structured correlation patterns of intact neighborhoods and the high-entropy, chaotic characteristics of debris structures and building damages.

\begin{figure*}[!ht]
\centering
\centering
\includegraphics[width=\linewidth]{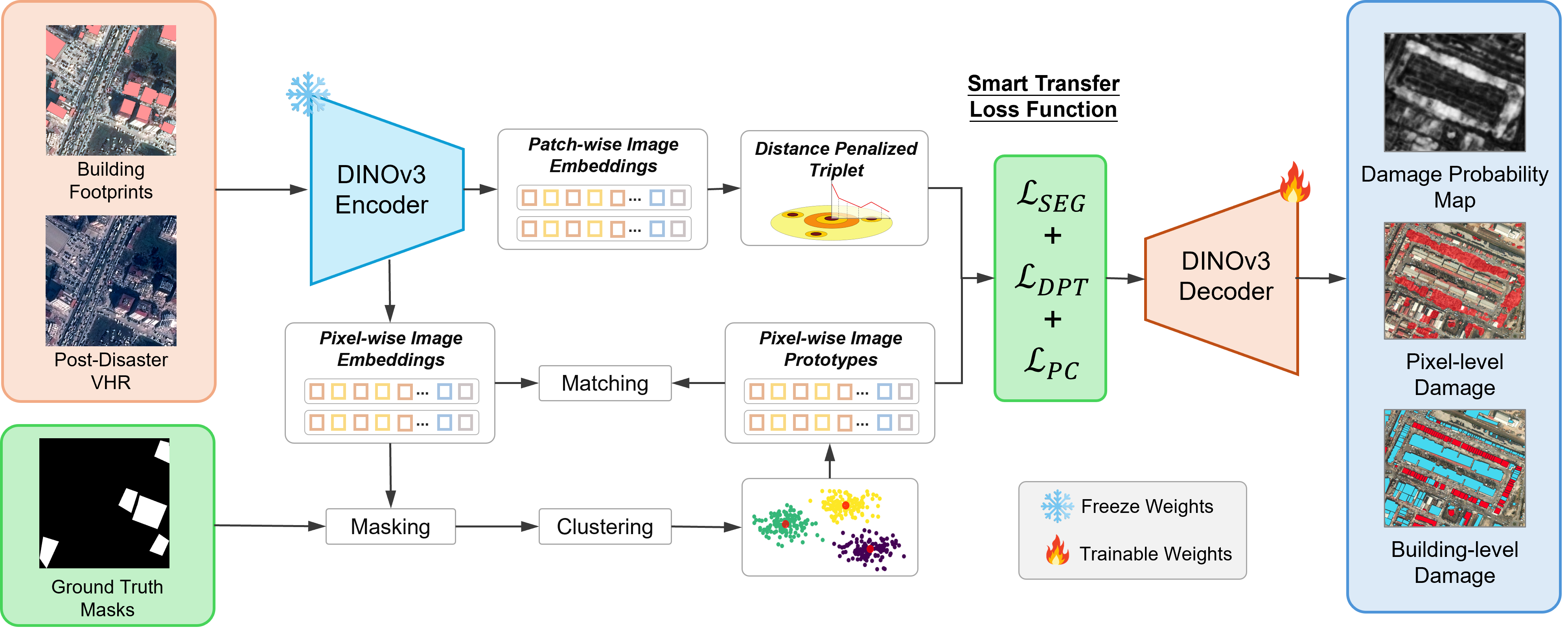}
\caption{Overall network architecture of Smart Transfer, consisting mainly of three parts. (1) the Freeze Vision FM Encoder, (2) Smart Transfer Loss Functions, (3) the Trainable Damage Mapping Decoder.}
\label{fig:architecture}
\end{figure*}

\subsection{Foundation Model Smart Transfer}

\textbf{PC} is designed to enhance smart transfer by regularizing dense pixel embeddings with stable class-conditional prototypes. 

Given a post-earthquake image patch $x$, we extract pixel-wise embeddings from the FM encoder and reshape them into a dense feature map $\mathbf{F}\in\mathbb{R}^{H\times W\times D}$. Using source-domain training data, pixel embeddings are separated according to ground-truth damage labels into damaged and non-damaged groups. Mini-batch $k$-means is independently applied to each group to construct two prototype sets, denoted as $\mathcal{P}^{+}$ and $\mathcal{P}^{-}$. These prototypes summarize representative visual patterns of each class (e.g., debris-like textures versus intact roof surfaces) and serve as stable reference anchors.

During optimization, each pixel embedding $\mathbf{f}(u,v)$ is assigned a prototype-derived pseudo-label via nearest-centroid matching over $\mathcal{P}^{+}\cup\mathcal{P}^{-}$. The identity of the closest prototype determines whether the pixel is associated with damaged or non-damaged regions. To prevent noisy prototype assignments from degrading optimization, we introduce a reliability constraint: only pixels whose prototype-derived labels agree with ground truth are used to compute the auxiliary loss. The PC objective is formulated as a masked binary cross-entropy between predicted probabilities and prototype-derived pseudo-labels:
\begin{equation}
\mathcal{L}_{PC}=
\frac{1}{|\Omega|}
\sum_{(u,v)\in\Omega}
\mathrm{BCE}\big(p(u,v),\,y^{proto}(u,v)\big),
\end{equation}
\begin{equation}
\Omega=\{(u,v)\mid y^{proto}(u,v)=\mathbf{Y}^{gt}(u,v)\},
\end{equation}
where $p(u,v)$ denotes the probability of predicted damage at the location $(u,v)$ and $\Omega$ is the set of reliable pixels.

This PC loss, as shown in Algorithm \ref{alg:pc_loss}, enforces within-cluster compactness and inter-cluster separability for different types of building damages, which significantly helps align the decoder's embedding space with the underlying morphological characteristics of building damages.

\textbf{DPT} serves as the second auxiliary loss in the smart transfer framework. Unlike PC, which regularizes pixel embeddings via class prototypes, DPT operates at the patch level and further regularizes the decoder by incorporating spatial autocorrelation patterns (i.e., damaged buildings and structural failures) into the embedding space, which is built upon our previous work in \cite{li2024gimi}.

Given decoder feature maps $\mathbf{F}\in\mathbb{R}^{H\times W\times D}$, each tile is partitioned into a regular $g\times g$ grid of non-overlapping patches. For each patch, spatial average pooling is applied to obtain a patch-level embedding, resulting in a set of patch features $\{\mathbf{e}_k\}_{k=1}^{g^2}$.

Patch labels are determined from the building footprint and the ground-truth damage mask. 
For each patch $k$, let $r_b$ denote the building coverage ratio and $r_d$ the damage ratio within building pixels. 
The patch label $y_k \in \{0,1,2,-1\}$ is assigned as:
\begin{equation}
y_k =
\begin{cases}
2, & r_b < \tau_b \\
1, & r_b \ge \tau_b \ \text{and} \ r_d \ge \tau_d \\
0, & r_b \ge \tau_b \ \text{and} \ r_d \le \tau_u \\
-1, & \text{otherwise}
\end{cases}
\end{equation}
where $\tau_b$ controls minimum building coverage, and $\tau_u,\tau_d$ are the lower and upper thresholds for confident non-damaged and damaged building patches.

Only confident building patches ($y_k\in\{0,1\}$) are used as anchor candidates. For each anchor $\mathbf{e}_a$, a positive $\mathbf{e}_p$ is sampled such that $y_p=y_a$, and a negative $\mathbf{e}_n$ is sampled such that $y_n\neq y_a$. The DPT objective is defined as:
\begin{equation}
\mathcal{L}_{DPT} = \max\Big(0,\,
\|\mathbf{e}_a-\mathbf{e}_p\|_2^2
-\|\mathbf{e}_a-\mathbf{e}_n\|_2^2
+P_{apn}
+\alpha\Big),
\end{equation}
\begin{equation}
P_{apn} = \frac{q_{ap}+q_{an}-q_{pn}}{\sqrt{2}},
\end{equation}
where $P_{apn}$ encodes the geometric relationship among the triplet, with $q_{ij}$ denoting the normalized Euclidean distance between patch centers $i$ and $j$. The parameter $\alpha$ is the margin. The pseudo code of DPT is detailed in Algorithm \ref{alg:dpt_loss}.

This formulation assigns stronger penalties to semantically inconsistent patches that are spatially adjacent, while relaxing constraints for distant regions. Since earthquake-induced damages typically exhibit strong spatial clustering, DPT enhances boundary discrimination between adjacent but semantically distinct patches (e.g., debris versus intact structures), thereby improving embedding separability and cross-domain robustness.

\begin{algorithm}[!ht]
\caption{PC}
\label{alg:pc_loss}
\begin{algorithmic}[1]
\Require Pixel embeddings $\mathbf{F} \in \mathbb{R}^{H \times W \times D}$, Ground-truth mask $\mathbf{Y}^{gt}$, Prototype sets $\mathcal{P}^{+}, \mathcal{P}^{-}$
\Ensure Auxiliary loss $\mathcal{L}_{PC}$
\Statex

\State \textbf{Step 1: Prototype Construction}
\State $\mathcal{Z}^{+} \leftarrow \emptyset,\;\mathcal{Z}^{-} \leftarrow \emptyset$
\For{$(x,\mathbf{Y}^{gt}) \in \mathcal{D}_s$}
    \State $\mathbf{F} \leftarrow \phi(x)$
    \State $\mathcal{Z}^{+} \leftarrow \mathcal{Z}^{+} \cup \{\mathbf{f}_i \mid Y^{gt}_i=1\}$
    \State $\mathcal{Z}^{-} \leftarrow \mathcal{Z}^{-} \cup \{\mathbf{f}_i \mid Y^{gt}_i=0\}$
\EndFor
\State $\mathcal{P}^{+} \leftarrow \mathrm{KMeans}(\mathcal{Z}^{+},K^{+})$
\State $\mathcal{P}^{-} \leftarrow \mathrm{KMeans}(\mathcal{Z}^{-},K^{-})$

\State \textbf{Step 2: Prototype Assignment}
\State For each pixel embedding $\mathbf{f}(u,v)$
\State \quad $k^{*} \leftarrow \arg\min_{\mathbf{p}_k \in \mathcal{P}^{+}\cup\mathcal{P}^{-}} \|\mathbf{f}(u,v) - \mathbf{p}_k\|_2^2$
\State \quad $y^{proto}(u,v) \leftarrow 
\begin{cases}
1, & k^{*} \le K^{+} \\
0, & \text{otherwise}
\end{cases}$

\State \textbf{Step 3: Reliability Gating}
\State $\Omega \leftarrow \{(u,v)\mid y^{proto}(u,v) = \mathbf{Y}^{gt}(u,v)\}$

\State \textbf{Step 4: Loss Computation}
\State $\mathcal{L}_{PC} \leftarrow 
\dfrac{1}{|\Omega|}
\sum_{(u,v)\in\Omega}
\mathrm{BCE}(p(u,v), y^{proto}(u,v))$

\State \Return $\mathcal{L}_{PC}$
\end{algorithmic}
\end{algorithm}

\subsection{Evaluation Metrics and Implementations}

\textbf{Evaluation Metrics:} In the context of the Smart Transfer approach, we evaluate building damage mapping performance from both building-level classification and pixel-level segmentation, respectively. To quantify the performance to classify building damages, we employ standard binary classification metrics, namely Precision, Recall, Accuracy, and the $F_1$-score. Precision and Recall are critical to balancing the trade-off between minimizing false alarms and ensuring that no critical damage is overlooked. The $F_1$-score serves as the harmonic mean of these two metrics, providing a robust measure of reliability in the presence of the class imbalance common in post-disaster satellite imagery.

Beyond building-level classification, we extend the evaluation to pixel-level segmentation performance between predicted masks and ground truth via the Intersection over Union (IoU) metric. We report $\text{IoU}_0$ (i.e., representing the background and undamaged class) and $\text{IoU}_1$ (i.e., representing the damaged class) to evaluate damage segmentation performance. Given that damaged structures often represent a small fraction of the total area, the Mean Intersection over Union (mIoU) is used as the primary benchmark for segmentation quality. This metric averages the IoU across both classes, ensuring that the model's performance on the minority ``damaged'' class is accurately represented towards a robust evaluation.

\textbf{Experiment Implementations:} All experiments were conducted on a single NVIDIA RTX 4090 GPU (24GB memory) and implemented in PyTorch. We adopt the DINOv3 ViT-L/16 encoder as the backbone, initialized with the publicly released checkpoint \footnote{dinov3\_vitl16\_pretrain\_sat493m-eadcf0ff}. A lightweight convolutional upsampling decoder (as shown in Figure~\ref{fig:architecture}) is attached for binary damage segmentation and trained from scratch.

\begin{algorithm}[ht!]
\caption{DPT}
\label{alg:dpt_loss}
\begin{algorithmic}[1]
\Require Patch embeddings $\mathbf{E}\in\mathbb{R}^{g^2\times D}$,
Patch labels $\mathbf{y}\in\{0,1,2,-1\}^{g^2}$,
Grid size $g$, Margin $\alpha$,
\Ensure $\mathcal{L}_{DPT}$
\Statex

\State \textbf{Step 1: Anchor Set}
\State $\mathcal{A} \leftarrow \{k \mid y_k \in \{0,1\}\}$

\State \textbf{Step 2: Triplet Sampling}
\State Shuffle $\mathcal{A}$
\For{$a \in \mathcal{A}$}
    \State $\mathcal{P} \leftarrow \{k \mid y_k = y_a,\; k \neq a\}$
    \State $\mathcal{N} \leftarrow \{k \mid y_k \neq y_a\}$
    \State Sample $p \sim \mathcal{P}$,\; $n \sim \mathcal{N}$
\EndFor

\State \textbf{Step 3: Spatial Penalty}
\State Compute patch centers $\mathbf{x}_k$
\State $q_{ap} \leftarrow \|\mathbf{x}_a-\mathbf{x}_p\|_2$
\State $q_{an} \leftarrow \|\mathbf{x}_a-\mathbf{x}_n\|_2$
\State $q_{pn} \leftarrow \|\mathbf{x}_p-\mathbf{x}_n\|_2$
\State $P_{apn} \leftarrow (q_{ap}+q_{an}-q_{pn})/\sqrt{2}$

\State \textbf{Step 4:  Loss Computation}
\State $\mathcal{L}_{DPT} \leftarrow
\max\big(0,
\|\mathbf{e}_a-\mathbf{e}_p\|_2^2
-\|\mathbf{e}_a-\mathbf{e}_n\|_2^2
+P_{apn}
+\alpha\big)$

\State \Return $\mathcal{L}_{DPT}$
\end{algorithmic}
\end{algorithm}

The input image resolution is $128\times128$. Optimization is performed using AdamW with an initial learning rate of $3\times10^{-4}$ and weight decay of $1\times10^{-4}$. A cosine learning rate schedule is applied, and gradient clipping is set to $1.0$. The batch size is $32$, and training runs for $10$ epochs per fold. 
Unless otherwise specified, results are averaged over five stratified folds. For few-shot adaptation, LoRA modules are injected into the encoder, while the backbone weights remain frozen. Only the LoRA parameters are optimized using a learning rate of $3\times10^{-5}$ for $3$ epochs, enabling parameter-efficient domain adaptation. To better understand the models’ behavior beyond performance metrics, we explore the internal attention patterns of Smart Transfer using Class Activation Maps (CAMs) \citep{selvaraju2017grad}.

\begin{table*}[!ht]
\centering
\caption{Full-supervision overall performance for building damage assessment, evaluating both classification and segmentation metrics (from five stratified folds). FMs indicate whether the model is a Foundation Model 
(\textcolor{green!70!black}{\ding{51}}) or not (\textcolor{red}{\ding{55}}).}
\label{tab:unified_full_supervision_fms}
\begin{tabular}{lc *{7}{r@{\,$\pm$\,}l}}
\toprule
Model & FMs & \multicolumn{2}{c}{Precision} & \multicolumn{2}{c}{Recall} & \multicolumn{2}{c}{Accuracy} & \multicolumn{2}{c}{F1} & \multicolumn{2}{c}{IoU$_0$} & \multicolumn{2}{c}{IoU$_1$} & \multicolumn{2}{c}{mIoU} \\
\midrule
ResNet-18    & \textcolor{red}{\ding{55}} & 0.14 & 0.06 & 0.68 & 0.07 & 0.44 & 0.03 & 0.22 & 0.08 & 0.38 & 0.05 & 0.12 & 0.05 & 0.25 & 0.02 \\
ResNet-152   & \textcolor{red}{\ding{55}} & 0.22 & 0.09 & 0.57 & 0.10 & 0.70 & 0.02 & 0.30 & 0.10 & 0.68 & 0.04 & 0.18 & 0.07 & 0.43 & 0.02 \\
YOLO-like   & \textcolor{red}{\ding{55}} & 0.23 & 0.10 & 0.55 & 0.15 & 0.73 & 0.06 & 0.31 & 0.11 & 0.71 & 0.07 & 0.19 & 0.08 & 0.45 & 0.03 \\
DINOv3       & \textcolor{green!70!black}{\ding{51}} & \textbf{0.55} & \textbf{0.09} & \textbf{0.75} & \textbf{0.08} & \textbf{0.90} & \textbf{0.02} & \textbf{0.62} & \textbf{0.09} & \textbf{0.89} & \textbf{0.03} & \textbf{0.46} & \textbf{0.09} & \textbf{0.68} & \textbf{0.04} \\
\bottomrule
\end{tabular}
\end{table*}

\begin{table*}[!ht]
\centering
\caption{Region-wise results for mIoU and F1 scores under full supervision (from five stratified folds). FMs indicate whether the model is a Foundation Model (\textcolor{green!70!black}{\ding{51}}) or not (\textcolor{red}{\ding{55}}).}
\label{tab:region_full_supervision_fms}
\resizebox{\textwidth}{!}{%
\begin{tabular}{l @{\hspace{1em}} *{4}{r@{\,$\pm$\,}l} @{\hspace{1.5em}} *{4}{r@{\,$\pm$\,}l}}
\toprule
 & \multicolumn{8}{c}{\textbf{mIoU}} & \multicolumn{8}{c}{\textbf{F1 Score}} \\
\cmidrule(lr){2-9} \cmidrule(lr){10-17}
 & \multicolumn{2}{c}{DINO} & \multicolumn{2}{c}{ResNet-18} & \multicolumn{2}{c}{ResNet-152} & \multicolumn{2}{c}{YOLO-like} & \multicolumn{2}{c}{DINO} & \multicolumn{2}{c}{ResNet-18} & \multicolumn{2}{c}{ResNet-152} & \multicolumn{2}{c}{YOLO-like} \\
 \midrule
FMs & \multicolumn{2}{c}{\textcolor{green!70!black}{\ding{51}}} & \multicolumn{2}{c}{\textcolor{red}{\ding{55}}} & \multicolumn{2}{c}{\textcolor{red}{\ding{55}}} & \multicolumn{2}{c}{\textcolor{red}{\ding{55}}} & \multicolumn{2}{c}{\textcolor{green!70!black}{\ding{51}}} & \multicolumn{2}{c}{\textcolor{red}{\ding{55}}} & \multicolumn{2}{c}{\textcolor{red}{\ding{55}}} & \multicolumn{2}{c}{\textcolor{red}{\ding{55}}} \\
\midrule
Gaziantep     & \textbf{0.61} & \textbf{0.05} & 0.25 & 0.06 & 0.41 & 0.05 & 0.45 & 0.06 & \textbf{0.47} & \textbf{0.08} & 0.11 & 0.03 & 0.16 & 0.04 & 0.19 & 0.08 \\
Hatay         & \textbf{0.63} & \textbf{0.03} & 0.26 & 0.07 & 0.41 & 0.06 & 0.44 & 0.03 & \textbf{0.55} & \textbf{0.05} & 0.17 & 0.01 & 0.24 & 0.04 & 0.21 & 0.02 \\
Kahramanmaras & \textbf{0.72} & \textbf{0.02} & 0.27 & 0.08 & 0.47 & 0.08 & 0.42 & 0.02 & \textbf{0.74} & \textbf{0.02} & 0.35 & 0.03 & 0.48 & 0.07 & 0.44 & 0.03 \\
Kirikhan      & \textbf{0.68} & \textbf{0.03} & 0.27 & 0.08 & 0.42 & 0.06 & 0.44 & 0.05 & \textbf{0.59} & \textbf{0.04} & 0.14 & 0.05 & 0.20 & 0.04 & 0.18 & 0.08 \\
Nurdagi       & \textbf{0.68} & \textbf{0.02} & 0.25 & 0.09 & 0.43 & 0.05 & 0.45 & 0.06 & \textbf{0.65} & \textbf{0.03} & 0.24 & 0.02 & 0.34 & 0.04 & 0.36 & 0.05 \\
Sakcagozu    & \textbf{0.65} & \textbf{0.03} & 0.22 & 0.08 & 0.41 & 0.06 & 0.40 & 0.08 & \textbf{0.60} & \textbf{0.06} & 0.21 & 0.03 & 0.31 & 0.05 & 0.30 & 0.04 \\
Satirhuyuk    & \textbf{0.70} & \textbf{0.02} & 0.24 & 0.09 & 0.45 & 0.07 & 0.50 & 0.05 & \textbf{0.68} & \textbf{0.02} & 0.24 & 0.02 & 0.35 & 0.06 & 0.42 & 0.05 \\
Sekeroba      & \textbf{0.67} & \textbf{0.03} & 0.24 & 0.08 & 0.41 & 0.05 & 0.45 & 0.04 & \textbf{0.58} & \textbf{0.05} & 0.16 & 0.01 & 0.22 & 0.04 & 0.25 & 0.06 \\
Turkoglu      & \textbf{0.73} & \textbf{0.07} & 0.28 & 0.09 & 0.46 & 0.04 & 0.51 & 0.06 & \textbf{0.75} & \textbf{0.08} & 0.35 & 0.04 & 0.42 & 0.07 & 0.49 & 0.08 \\
\midrule
Overall       & \textbf{0.68} & \textbf{0.04} & 0.25 & 0.02 & 0.43 & 0.02 & 0.45 & 0.03 & \textbf{0.62} & \textbf{0.09} & 0.22 & 0.08 & 0.30 & 0.10 & 0.31 & 0.11 \\
\bottomrule
\end{tabular}
}
\end{table*}

During training, the segmentation loss is implemented as focal loss to mitigate foreground sparsity, with parameters $\alpha=0.7$ and $\gamma=2.0$. For PC, the numbers of positive and negative prototypes are both set to $32$, and prototype matching is performed in the embedding space. For DPT, each tile is partitioned into a $4\times4$ grid. The margin is set to $\alpha=0.2$. Patch labels are assigned using thresholds $\tau_b=0.02$, $\tau_u=0.02$, and $\tau_d=0.10$. Spatial distances are normalized by the maximum patch-center distance within each tile. The overall Smart Transfer training objective is defined as:
\begin{equation}
\mathcal{L_{ST}} = \mathcal{L}_{SEG}
+ \lambda_{pc}\mathcal{L}_{PC}
+ \lambda_{dpt}\mathcal{L}_{DPT},
\end{equation}
where the auxiliary loss weights are set to $\lambda_{pc}=0.01$ and $\lambda_{dpt}=0.001$, selected via ablation studies.

During inference, a sigmoid activation is applied to generate probability maps, and binary masks are obtained by thresholding at $0.5$. Building-level metrics are computed by aggregating pixel predictions within building footprints, while segmentation performance is evaluated at the pixel level.

\section{Experiment} \label{experiment}

\subsection{Fully Supervised Baseline}

To evaluate the proposed Smart Transfer approach, the first step is to establish a Fully Supervised (FS) competitive baseline. Herein, we consider three non-FM baselines to compare with the DINOv3 vision FM \citep{simeoni2025dinov3}, specifically, two ResNets \citep{he2016deep} with 18 and 152 layers as well as a YOLO-like model \citep{redmon2016you}. Herein, we first present the FS results, as they set a competitive baseline for all Smart Transfer models to beat, based on the common assumption that the FS method provides a strong baseline for cross-regional damage mapping.

Table \ref{tab:unified_full_supervision_fms} shows the quantitative comparison of these FS baselines from both damage segmentation and building-level classification perspectives. As one can observe, under the full-supervision setting, the FM-based architecture (DINOv3) demonstrates superior accuracy performance in both tasks. Specifically, the DINOv3 FM consistently outperforms the non-FM alternatives, achieving the highest overall F1-score (0.62) and mean Intersection over Union (mIoU) (0.68). Moreover, DINOv3 also exhibits the strongest recall (0.75), highlighting its superior capacity to detect structural damage while retaining competitive precision. Among the non-FM baselines, ResNet-152 and the YOLO-like model exhibit comparable overall performance, with the YOLO-like architecture achieving slightly better segmentation metrics in most cases. In contrast, ResNet-18 shows a tendency toward over-prediction, characterized by relatively high recall but substantially lower precision, leading to inferior overall performance.

Table \ref{tab:region_full_supervision_fms} further provides region-wise performance metrics (i.e., F1 for building-level classification and mIoU for damage segmentation), which confirms the key findings in Table \ref{tab:unified_full_supervision_fms} about DINOv3's performance, as it secures the highest mIoU across all geographical areas. Though ResNet-152 and the YOLO-like model perform competitively in specific regions such as Kahramanmaras, the transformer-based vision FM delivers far more stable and consistent generalization across diverse urban morphologies. Ultimately, these findings empirically validate that high-capacity, pre-trained vision encoders drastically enhance building damage segmentation compared to traditional convolutional networks.

Overall, the FS results highlight the superior performance of vision FMs, specifically DINOv3, in both damage segmentation and building-level classification, providing a solid basis for further performance improvements to Smart Transfer to enhance the geographical generalizability and data-driven adaptation of vision FMs in time-critical disaster response scenarios.


\begin{table*}[t!]
\centering
\caption{Performance w.r.t mIoU and F1 scores for building damage assessment under the LODO setting (from five stratified folds). FS refers to the fully supervised DINOv3 model, which serves as a competitive baseline.}
\label{tab:unified_lodo_results}
\resizebox{\textwidth}{!}{%
\begin{tabular}{l @{\hspace{1em}} *{4}{r@{\,$\pm$\,}l} @{\hspace{1.5em}} *{4}{r@{\,$\pm$\,}l}}
\toprule
 & \multicolumn{8}{c}{\textbf{mIoU}} & \multicolumn{8}{c}{\textbf{F1 Score}} \\
\cmidrule(lr){2-9} \cmidrule(lr){10-17}
Target Domain & \multicolumn{2}{c}{FS} & \multicolumn{2}{c}{PC} & \multicolumn{2}{c}{DPT} & \multicolumn{2}{c}{PC+DPT} & \multicolumn{2}{c}{FS} & \multicolumn{2}{c}{PC} & \multicolumn{2}{c}{DPT} & \multicolumn{2}{c}{PC+DPT} \\
\midrule
Gaziantep     & 0.61 & 0.05 & 0.65 & 0.04 & 0.65 & 0.03 & \textbf{0.66} & \textbf{0.03} & 0.47 & 0.08 & 0.51 & 0.07 & 0.52 & 0.06 & \textbf{0.53} & \textbf{0.05} \\
Hatay         & 0.63 & 0.03 & \textbf{0.64} & \textbf{0.03} & 0.63 & 0.05 & 0.63 & 0.03 & 0.55 & 0.05 & \textbf{0.56} & \textbf{0.05} & 0.55 & 0.07 & 0.55 & 0.05 \\
Kahramanmaras & \textbf{0.72} & \textbf{0.02} & 0.69 & 0.03 & 0.68 & 0.05 & 0.67 & 0.03 & \textbf{0.74} & \textbf{0.02} & 0.67 & 0.05 & 0.64 & 0.07 & 0.63 & 0.06 \\
Kirikhan      & 0.68 & 0.03 & 0.70 & 0.07 & \textbf{0.71} & \textbf{0.04} & 0.70 & 0.07 & 0.59 & 0.04 & 0.62 & 0.11 & \textbf{0.64} & \textbf{0.05 }& 0.62 & 0.11 \\
Nurdagi       & 0.68 & 0.02 & \textbf{0.71} & \textbf{0.02} & 0.70 & 0.03 & 0.70 & 0.03 & 0.65 & 0.03 & \textbf{0.69} & \textbf{0.03} & 0.68 & 0.03 & 0.68 & 0.04 \\
Sakcagozu    & 0.65 & 0.03 & \textbf{0.72} & \textbf{0.04} & 0.71 & 0.04 & 0.72 & 0.05 & 0.60 & 0.06 & \textbf{0.68} & \textbf{0.06} & 0.67 & 0.05 & 0.68 & 0.07 \\
Satirhuyuk    & 0.70 & 0.02 & \textbf{0.74} & \textbf{0.02} & 0.73 & 0.02 & 0.74 & 0.02 & 0.68 & 0.02 & 0.71 & 0.03 & 0.71 & 0.03 & \textbf{0.71} & \textbf{0.03} \\
Sekeroba      & 0.67 & 0.03 & \textbf{0.68} & \textbf{0.03} & 0.65 & 0.06 & 0.66 & 0.04 & 0.58 & 0.05 & \textbf{0.58} & \textbf{0.06} & 0.52 & 0.14 & 0.56 & 0.07 \\
Turkoglu      & \textbf{0.73} & \textbf{0.07} & 0.72 & 0.06 & 0.72 & 0.03 & 0.72 & 0.06 & \textbf{0.75} & \textbf{0.08} & 0.71 & 0.08 & 0.72 & 0.05 & 0.71 & 0.09 \\
\midrule
Overall       & 0.68 & 0.04 & \textbf{0.69} & \textbf{0.03} & 0.69 & 0.03 & 0.69 & 0.03 & 0.62 & 0.09 & \textbf{0.64} & \textbf{0.06} & 0.63 & 0.06 & 0.63 & 0.06 \\
\bottomrule
\end{tabular}%
}
\end{table*}

\begin{figure*}[t!]
\centering
\centering
\includegraphics[width=\linewidth]{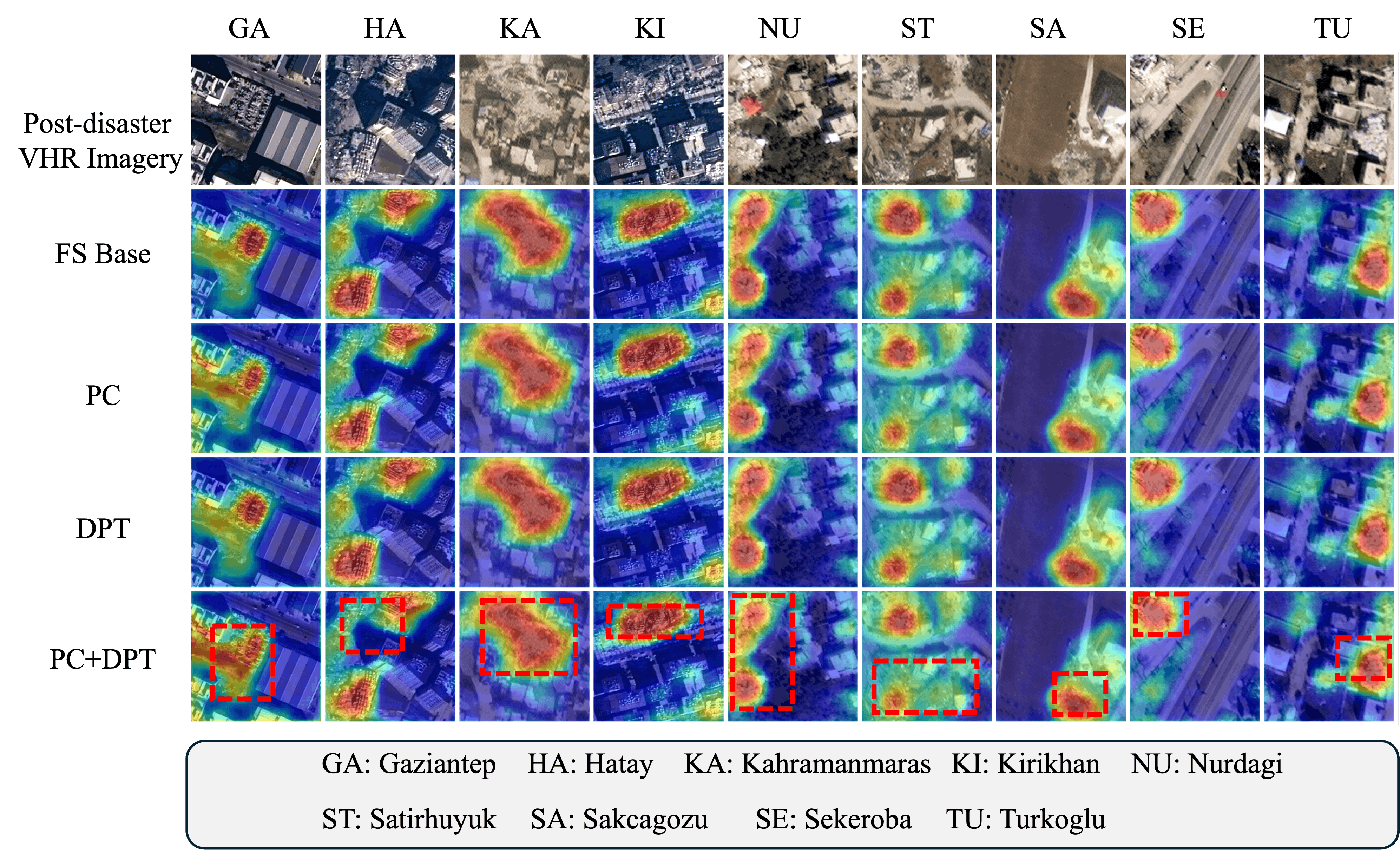}
\caption{Visualization of the regional-wise Class Activation Maps (CAMs) of the fully supervised baselines compared to different smart transfer variants.}
\label{fig:heatmap_PCDPT}
\end{figure*}

\subsection{Performance Improvements of Smart Transfer}
\textbf{Transfer Setting 1 (LODO):} The LODO domain generalization setting serves as a rigorous benchmark to evaluate the proposed Smart Transfer approach by strictly isolating target domains from the training corpus. By aggregating training signals from multiple source domains while withholding one geographically distinct target domain, LODO systematically tests the model's capacity to extrapolate learned structural knowledge and generalize across unseen, diverse urban morphologies. To facilitate this cross-regional transfer, we design two Smart Transfer strategies, namely PC and DPT. Specifically, PC regularizes dense pixel-wise embeddings using clustering-based prototypes. This enforces within-cluster compactness and inter-cluster separability, aligning the decoder's embedding space with the fundamental morphological characteristics of building damages. Additionally, DPT works at the patch level, explicitly incorporating spatial autocorrelation patterns by assigning stronger penalties to semantically inconsistent yet spatially adjacent patches. This explicitly enhances boundary discrimination between structural debris and intact edifices, improving embedding separability.

Table \ref{tab:unified_lodo_results} details experimental results under the LODO setting, indicating that PC yields the most robust cross-domain generalization performance. Overall, PC achieves the best accuracy in both damage segmentation and building-level classification with a mean mIoU of 0.69 ± 0.03 and an F1 score of 0.64 ± 0.06. Region-wise, LODO further corroborates this finding, with PC achieving the highest mIoU across most target domains, including Hatay, Nurdagi, Sakcagozu, Satirhuyuk and Sekeroba. However, DPT and PC+DPT show highly localized efficiencies rather than cross-region performance gains. 

Figure \ref{fig:heatmap_PCDPT} presents the regional-wise CAM responses of the FS baseline as well as different Smart Transfer Variations. One can clearly notice that Smart Transfer achieves competitive attention activation of structural building damages across regions, almost similar to the FS, which represents the fully supervised baseline. 

Importantly, the LODO results shed light on the Smart Transfer of FMs for geographically diverse damage mapping. Prototype-level alignment through PC inherently shows more stable generalization performance by maintaining global feature alignment and adapting to the distinct urban morphology. Herein, this is also why the patch-level spatial regularization via DPT does not necessarily lead to performance boosting, as the LODO is addressed better via PC already. In summary, Smart Transfer via PC ensures explicit spatial priors can be embedded into the FMs as topological diversity is already implicitly maximized within the training corpus, rendering PC the dominant strategy for cross-domain transfer in data-rich generalization scenarios.

\begin{table*}[t!]
\centering
\caption{Performance w.r.t mIoU and F1 scores for building damage assessment under the SSDC setting (from five stratified folds). FS refers to the fully supervised DINOv3 model, which serves as a competitive baseline.}
\label{tab:ssdc_combined}
\scriptsize
\begin{tabularx}{\textwidth}{l @{\extracolsep{\fill}} cccc cccc}
\toprule
& \multicolumn{4}{c}{\textbf{mIoU}} & \multicolumn{4}{c}{\textbf{F1 Score}} \\
\cmidrule(lr){2-5} \cmidrule(lr){6-9}
Source Domain & FS & PC & DPT & PC+DPT & FS & PC & DPT & PC+DPT \\
\midrule
Kah           & 0.68 ± 0.04 & 0.68 ± 0.06 & 0.68 ± 0.06 & \textbf{0.67 ± 0.06} & 0.62 ± 0.09 & 0.62 ± 0.11 & \textbf{0.63 ± 0.11} & 0.63 ± 0.11 \\
Nur           & 0.68 ± 0.04 & 0.68 ± 0.05 & \textbf{0.68 ± 0.04} & 0.68 ± 0.04 & 0.62 ± 0.09 & 0.59 ± 0.09 & \textbf{0.61 ± 0.07} & 0.60 ± 0.08 \\
Kah+Nur       & 0.68 ± 0.04 & \textbf{0.71 ± 0.04} & 0.70 ± 0.05 & 0.71 ± 0.04 & 0.62 ± 0.09 & \textbf{0.66 ± 0.09} & 0.64 ± 0.10 & 0.65 ± 0.09 \\
Kah+Hat       & 0.68 ± 0.04 & 0.70 ± 0.04 & \textbf{0.71 ± 0.04} & 0.70 ± 0.04 & 0.62 ± 0.09 & 0.65 ± 0.09 & \textbf{0.66 ± 0.08} & 0.64 ± 0.09 \\
Nur+Hat       & 0.68 ± 0.04 & \textbf{0.70 ± 0.03} & 0.69 ± 0.03 & 0.69 ± 0.03 & 0.62 ± 0.09 & \textbf{0.63 ± 0.06} & 0.62 ± 0.06 & 0.62 ± 0.07 \\
Kah+Nur+Hat   & 0.68 ± 0.04 & 0.70 ± 0.04 & \textbf{0.71 ± 0.03} & 0.70 ± 0.04 & 0.62 ± 0.09 & 0.64 ± 0.09 & \textbf{0.67 ± 0.07} & 0.64 ± 0.08 \\
\midrule
Overall       & 0.68 ± 0.00 & 0.69 ± 0.01 & \textbf{0.69 ± 0.01} & 0.69 ± 0.01 & 0.62 ± 0.00 & 0.63 ± 0.02 & \textbf{0.64 ± 0.02} & 0.63 ± 0.02 \\
\bottomrule
\end{tabularx}
\end{table*}

\begin{figure*}[t!]
     \centering
     \begin{subfigure}[b]{0.48\textwidth}
         \centering
         \includegraphics[width=\textwidth]{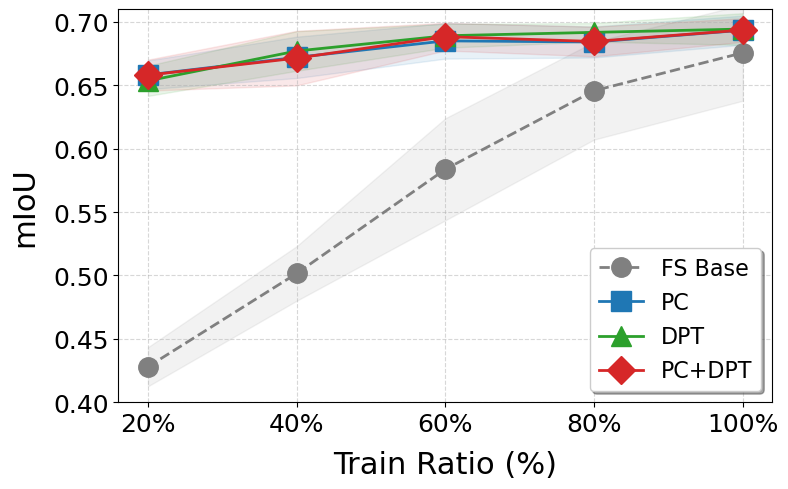}
         \label{fig:left}
     \end{subfigure}
     \hfill 
     \begin{subfigure}[b]{0.48\textwidth}
         \centering
         \includegraphics[width=\textwidth]{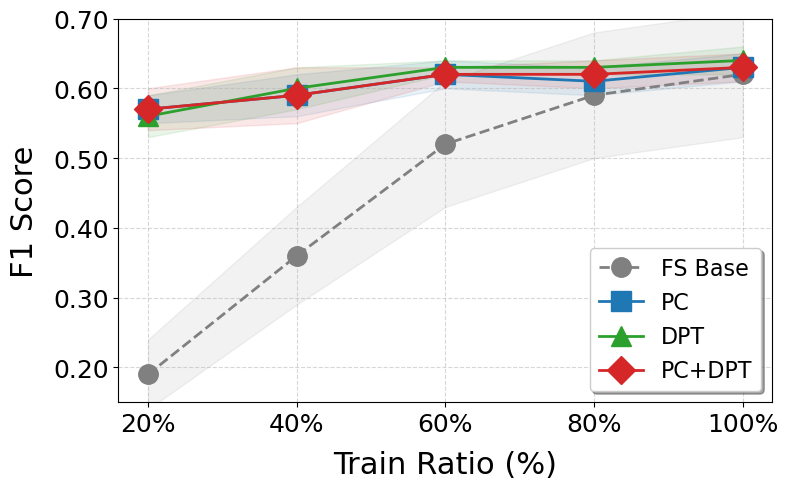}
         \label{fig:right}
     \end{subfigure}
     
     \caption{Influence of different train ratios on model performances (w.r.t mIoU and F1 scores) across the 6 selected source domains under the SSDC setting.}
     \label{fig:train_ratio_SSDC}
\end{figure*}

\textbf{Transfer Setting 2 (SSDC):} The SSDC setting significantly differs from the LODO paradigm by constraining source domain diversity. To ensure a fair comparison between different source-domain combinations, the SSDC results are computed as the macro average across all nine regions. Unlike implicitly maximizing spatial heterogeneity via LODO, SSDC is designed to isolate and evaluate the specific impact of source domain selection based on primarily two factors: 1) damage severity and 2) geographical distance.

\textbf{Damage Level Domain:} We first analyze the influence of damage severity in source-domain selection. Specifically, in Table \ref{tab:ssdc_combined}, Kahramanmaras (Kah) and Nurdagi (Nur) represent the two most severely affected regions in the dataset, characterized by higher proportions of damaged buildings. When trained on either Kah or Nur alone, the model achieves moderate generalization performance with baseline mean Intersection over Union (mIoU) scores of 0.67 ± 0.06 and 0.68 ± 0.04, respectively. Notably, combining Kah and Nur leads to a clear improvement in mIoU and F1 across all variants, suggesting that heavy-damage domains provide rich supervision signals for learning building damage representations.
More importantly, a key finding is that the DPT strategy becomes more effective when multiple heavy-damage domains are used as source domains. This finding indicates that pixel-level prototype via PC excels in LODO settings, leveraging structure damage similarities, whereas patch-level spatial regularization via DPT benefits from geographically clustered structural diversity within severely damaged regions.

\textbf{Geographical Distance Domain:} Moreover, we further examine the role of geographical distance by grouping source domains according to their locations. In this context, we identified three regions, namely Kahramanmaras (north), Nurdagi (central), and Hatay (south), together covering the full latitudinal span of the study area. As shown in Table \ref{tab:ssdc_combined}, incorporating geographically distant domains generally improves generalization performance, as evidenced by the highest mIoU $(0.71 \pm  0.03) $and F1 score $(0.67 \pm 0.07)$ achieved under the three-domain setting (Kah+Nur+Hat). Importantly, unlike in the LODO setting, DPT demonstrates stronger benefits when spatially diverse domains are combined. In the tri-domain Kah+Nur+Hat configuration, DPT attains an mIoU of $0.71 \pm  0.03$ and an F1 score of $0.67 \pm  0.07$. Consequently, DPT achieves the highest overall performance in the SSDC setting, with an mIoU of $0.69 \pm  0.01 $and an F1 score of $0.64 \pm  0.02$. This result suggests that spatial autocorrelation modeling becomes more effective when the training data captures heterogeneous structural patterns across geographically distant regions, compensating for the incomplete structural coverage inherent to limited source domain selections

In summary, these findings indicate that domain diversity should be characterized not only by regions but also by structural damage levels and spatial locations, which interact differently with pixel-level and patch-level regularization strategies.

\begin{table*}[htbp]
\centering
\scriptsize
\caption{Performance comparison of post-disaster building damage classification in Kahramanmaras under both LODO and SSDC settings. Raw refer to the non-adapted transfer.}
\label{tab:performance_metrics}
\begin{tabular}{lcccccccc}
\toprule
\multirow{2}{*}{Metric} & \multicolumn{4}{c}{LODO} & \multicolumn{4}{c}{SSDC} \\
\cmidrule(lr){2-5} \cmidrule(lr){6-9} 
 & Raw  & PC & DPT & PC+DPT & Raw  & PC & DPT & PC+DPT \\
\midrule
Smart Transfer & \textcolor{red}{\ding{55}} & \textcolor{green!70!black}{\ding{51}} & \textcolor{green!70!black}{\ding{51}} & \textcolor{green!70!black}{\ding{51}} & \textcolor{red}{\ding{55}} & \textcolor{green!70!black}{\ding{51}} & \textcolor{green!70!black}{\ding{51}} & \textcolor{green!70!black}{\ding{51}} \\
\midrule
Precision & \textbf{0.96} & 0.91 & 0.95 & 0.94 & 0.94 & 0.95 & 0.93 & 0.92 \\
Recall & 0.22 & \textbf{0.30} & 0.24 & 0.19 & 0.20 & 0.17 & 0.18 & 0.16 \\
Accuracy & 0.74 & \textbf{0.76} & 0.74 & 0.72 & 0.73 & 0.72 & 0.72 & 0.71 \\
F1 & 0.36 & \textbf{0.46} & 0.38 & 0.32 & 0.33 & 0.29 & 0.31 & 0.28 \\
IoU0 & 0.71 & \textbf{0.73} & 0.72 & 0.70 & 0.71 & 0.70 & 0.70 & 0.70 \\
IoU1 & 0.22 & \textbf{0.29} & 0.23 & 0.19 & 0.20 & 0.17 & 0.18 & 0.16 \\
mIoU & 0.47 & \textbf{0.51} & 0.48 & 0.45 & 0.45 & 0.44 & 0.44 & 0.43 \\
\bottomrule
\end{tabular}
\end{table*}

\begin{figure*}[t!]
    \centering
    \includegraphics[width=0.75\linewidth]{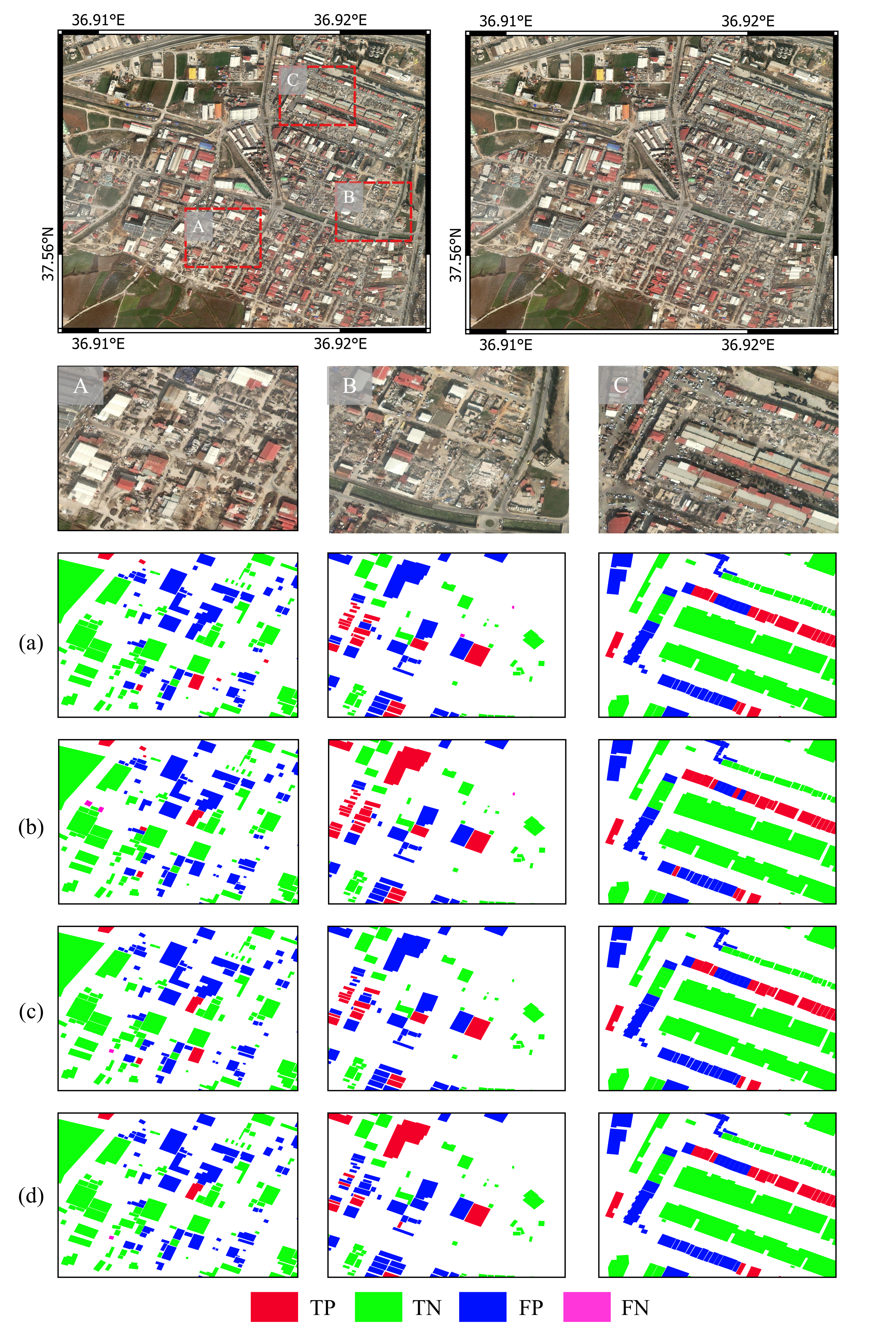}
    \caption{Post-disaster Building Damage Classification for a part of Kahramanmaras. (a) Raw, (b) PC, (c) DPT, (d) PC+DPT.}
    \label{fig:Infer_region}
\end{figure*}

\textbf{Training Ratio Impact:} Figure \ref{fig:train_ratio_SSDC} shows an ablation analysis on the impact of varying source-domain training ratios in the SSDC setting. Herein, the training ratio is a key factor in real-world disaster response. In case of an earthquake, the annotations of building damages are often collected step-by-step and region-by-region. Therefore, we aim to determine the exact training ratio sufficient to achieve reasonable performance for Smart Transfer under the SSDC setting. Noticeably, the FS baseline is highly sensitive to the amount of training data, with performance dropping substantially as the training ratio decreases. In contrast, the proposed Smart Transfer approaches demonstrate strong robustness under limited supervision. In the low-data regime (20-40\%), PC, DPT, and PC+DPT all achieve markedly higher mIoU and F1 scores than the FS baseline, indicating that the proposed transfer strategies can effectively compensate for insufficient source-domain annotations. As the training ratio increases, the FS baseline gradually improves and the gap becomes smaller, but the proposed methods remain consistently competitive across all data scales. These findings suggest that Smart Transfer significantly enhances data efficiency, enabling the FMs to extract more transferable structural knowledge from limited source-domain samples. This feature is particularly important for real-world post-disaster scenarios, where annotated data is collected with priorities, thus unevenly distributed across regions.

\subsection{Towards Building-level Damage Mapping and Deployment}


As demonstrated in the previous experiments, Smart Transfer achieves competitive performance compared to full supervision. Here, we further validate its practical applicability by comparing it with non-adapted transfer through building-level damage classification in a representative region of Kahramanmaras. Table \ref{tab:performance_metrics} summarizes the quantitative results under both LODO and SSDC settings, while Figure \ref{fig:Infer_region} presents the corresponding qualitative damage maps.

Under the LODO setting, Smart Transfer significantly improves the detection of damaged buildings. Compared with the Raw baseline, the PC strategy achieves the best overall performance with an F1-score of 0.46 and an mIoU of 0.51. The improvement mainly comes from the increase in recall, suggesting that prototype-based feature alignment effectively reduces missed detections of damaged structures. Although the Raw baseline achieves very high precision, its conservative prediction behavior leads to many false negatives. In contrast, PC provides a more balanced trade-off between precision and recall, leading to more reliable building-level damage classification.

Under the SSDC setting, we evaluate Smart Transfer using Nurdagi and Hatay (Nur+Hat) as the source-domain combination. Compared with the LODO setting,  overall performance is slightly lower due to reduced diversity of training domains. The Smart Transfer variants show comparable performance to the Raw baseline, but the improvements observed in LODO become less evident. This suggests that when the number of source domains is limited, the effectiveness of Smart Transfer is constrained because the Nur+Hat combination covers a narrower range of structural damage patterns and thus provides less transferable spatial knowledge.

Figure \ref{fig:Infer_region} further illustrates the qualitative differences among methods. The Raw baseline produces sparse damage predictions and misses many damaged buildings in dense urban areas. In contrast, the Smart Transfer variants capture more coherent clusters of damaged structures that better resemble the reference map. Among them, the PC-based model generates the most balanced prediction pattern, while DPT-based variants produce smoother spatial distributions due to spatial autocorrelation constraints. These results confirm that Smart Transfer improves the reliability of building-level damage mapping across different transfer settings.


\subsection{Ablation Study}

For Smart Transfer, two factors deserve additional attention: 1) balance and tradeoff between PC and DPT, as we have observed a distinct performance behavior in both LODO and SSDC settings; 2) robustness of FM encoders, as the Smart Transfer mainly uses a freeze encoder and trainable lightweight decoder; therefore, one might wonder if additional training of FMs encoders can also help. To partly answer these, we conducted two ablation studies.

\begin{table*}[!ht]
\centering
\caption{Few-shot learning results (mIoU) under both LODO and SSDC, (from five stratified folds).}
\begin{tabular}{llccccc}
\toprule
Setting & Shot & Finetune & Raw & PC & DPT & PC+DPT \\
\midrule
LODO  & 0-shot      & \textcolor{red}{\ding{55}} & 0.6851 ± 0.0370 & \textbf{0.6932 ± 0.0325} & 0.6876 ± 0.0345 & 0.6885 ± 0.0332 \\
      & 1-shot   & \textcolor{green!70!black}{\ding{51}} & 0.6865 ± 0.0358 & \textbf{0.6944 ± 0.0320} & 0.6891 ± 0.0335 & 0.6897 ± 0.0323 \\
      & 3-shot   & \textcolor{green!70!black}{\ding{51}} & 0.6867 ± 0.0364 & \textbf{0.6946 ± 0.0327} & 0.6892 ± 0.0344 & 0.6900 ± 0.0331 \\
      & 5-shot   & \textcolor{green!70!black}{\ding{51}} & 0.6868 ± 0.0361 & \textbf{0.6949 ± 0.0328} & 0.6895 ± 0.0341 & 0.6902 ± 0.0331 \\
      & 10-shot  & \textcolor{green!70!black}{\ding{51}} & 0.6878 ± 0.0358 & \textbf{0.6916 ± 0.0399} & 0.6902 ± 0.0338 & 0.6910 ± 0.0328 \\
\midrule
SSDC  & 0-shot      & \textcolor{red}{\ding{55}} & 0.6924 ± 0.0131 & 0.6933 ± 0.0115 & \textbf{0.6942 ± 0.0125} & 0.6933 ± 0.0094 \\
      & 1-shot   & \textcolor{green!70!black}{\ding{51}} & 0.6933 ± 0.0129 & 0.6940 ± 0.0114 & \textbf{0.6950 ± 0.0123} & 0.6941 ± 0.0093 \\
      & 3-shot   & \textcolor{green!70!black}{\ding{51}} & 0.6936 ± 0.0126 & 0.6943 ± 0.0113 & \textbf{0.6954 ± 0.0120} & 0.6943 ± 0.0092 \\
      & 5-shot   & \textcolor{green!70!black}{\ding{51}} & 0.6938 ± 0.0126 & 0.6945 ± 0.0112 & \textbf{0.6955 ± 0.0120} & 0.6945 ± 0.0091 \\
      & 10-shot  & \textcolor{green!70!black}{\ding{51}} & 0.6941 ± 0.0125 & 0.6949 ± 0.0112 & \textbf{0.6959 ± 0.0119} & 0.6949 ± 0.0092 \\
\bottomrule
\label{tab:few_short}
\end{tabular}
\end{table*}
\begin{figure*}[!ht]
\centering
\centering
\includegraphics[width=\linewidth]{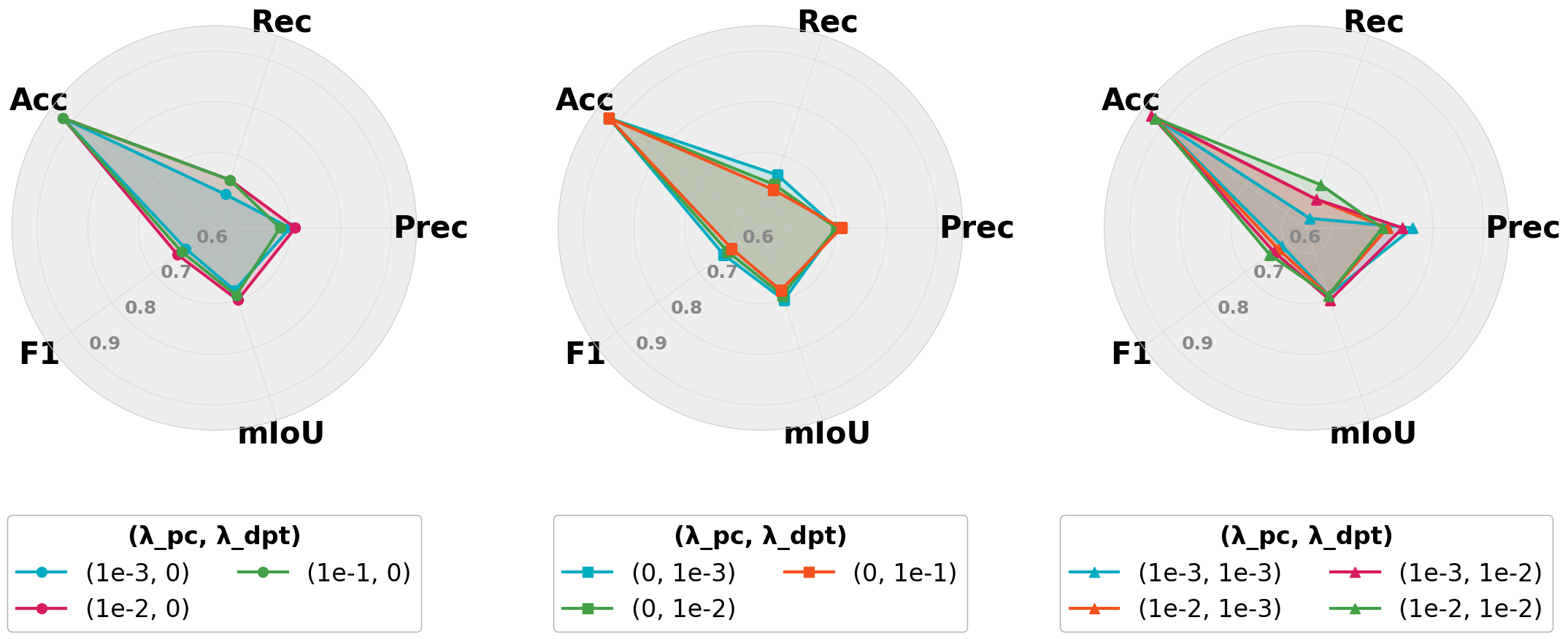}
\caption{Tradeoff between PC and DPT considering different weighting strategies. Herein, $\lambda_{pc}$ and $\lambda_{dpt}$ refer to the loss function weights of PC and DPT, respectively. }
\label{fig:PCDPT_tradeoff}
\end{figure*}
\textbf{Tradeoff between PC and DPT:} First, we analyze the sensitivity of Smart Transfer to the weighting coefficients of PC and DPT under the LODO setting. Figure \ref{fig:PCDPT_tradeoff} shows different tradeoff variables under three weightings (i.e., PC, DPT, and PC+DPT). On the one hand, PC achieves optimal representation alignment at $\lambda_{pc}=0.01$. On the other hand, excessive DPT constraints (e.g., $\lambda_{dpt}=0.1$) degrade mapping accuracy, indicating that overly aggressive patch-level spatial regulation distorts semantic alignment under domain shift. For PC and DPT jointly, the optimal configuration necessitates a balanced combination (e.g., $\lambda_{pc}=0.01,\lambda_{dpt}=0.001$) rather than independently maximized weights. This confirms that while Smart Transfer demonstrates robustness to minor hyperparameter variations, it is suggested to balance PC and DPT weights to prevent pixel-level prototype alignment and patch-level spatial constraints in real-world deployment.

\textbf{Few-shot Learning:} Second, we test the few-shot finetuning of DINOv3 encoders under both LODO and SSDC settings. Table \ref{tab:few_short} shows that the actual performance gain of few-shot learning is very limited. In the LODO setting, the model leverages high source-domain diversity to learn comprehensive spatial damage patterns. PC's superiority is confirmed across all k-shot settings in LODO, escalating from a frozen 0-shot baseline of $0.6932 \pm 0.0325$ to a 5-shot peak of $0.6949 \pm 0.0328$. In contrast, under SSDC settings, DPT becomes more effective. DPT outperforms under the SSDC setting, slightly improving from a 0-shot baseline of $0.6942  \pm 0.0125$ to $0.6959 \pm 0.0119$ under 10-shot fine-tuning; however, the improvement is still considered rather trivial.  Actually, this is what we want to show, as we believe the capability of vision FMs to act as general-purpose imagery encoders is well established, and the potential of Smart Transfer boosting lies mainly in the lightweight decoder.

\section{Discussions} \label{dicussion}
Despite the encouraging results of the proposed Smart Transfer approach, we identified some limitations in the current design and application, which we hope to share with the broader disaster response research community and encourage future work in this direction.

\textbf{Visual Language Model for Fine-grained Mapping:} In this work, we mainly rely on post-disaster VHR imagery to detect and map building damages, which largely limits the damage perception granularity and semantic context. In real-world disaster assessment, one often demands more fine-grained damage classifications, differentiating between slightly damaged buildings and structures requiring demolition. In this context, we argue that auxiliary geospatial data streams, such as ground-level Street View Imagery (SVI) \citep{biljecki2021street, li2025cross, liang2026heterogeneous} and crowdsourced social media content \citep{zou2018mining, hu2023geo, yin2025llm}, can supply key complementary information from either a cross-view or natural language perspective. 

Towards bridging multimodal geospatial information, Vision Language Models (VLMs) have shown substantial potential to leverage both language- and vision-based information for fine-grained damage mapping and disaster perception. Earlier works in this direction by \citet{wang2026cityvlm, yang2025perceiving, liang2026heterogeneous} highlight the potential of incorporating VLMs or LLMs in supporting fine-grained geospatial prediction and perception tasks in complex urban built environments, which can significantly benefit the rapid disaster mapping tasks as well in the future.

\textbf{Autonomous GIS Workflow in Disaster Response:} Though the Smart Transfer significantly accelerates the mapping of building damage from post-disaster VHR imagery, there are still significant technical barriers when it comes to transfer settings and workflow optimization. In time-critical humanitarian assistance and disaster response scenarios, isolated mapping capabilities are insufficient for immediate disaster response deployment. 

Moreover, to truly demonstrate rapid response capabilities, a cross-event evaluation will be essential. In a real-world disaster scenario, assuming the availability of any ground-truth labels for a new event is mostly unrealistic. The current experiments from this paper are preliminary, limited to intra-event transfer; therefore, the model's generalization may still significantly vary across different urban morphologies and disaster characteristics. 

Given the recent trend of autonomous GIS with GeoAI \citep{li2023autonomous, li2025giscience}, future works are encouraged to consider integrating Smart Transfer into an autonomous GIS workflow optimized in terms of disaster types, affected areas, and prioritized response missions. Another promising way is to develop an agent-based workflow to autonomously generate structured disaster reports, with which local stakeholders can directly optimize resource allocation during the critical ``Golden 72 Hours'' of search and rescue operations.

\textbf{Towards Disaster Resilience for Local Communities:} Herein, the Smart Transfer framework preliminarily focuses on post-disaster damage assessment, aiming at shortening the rapid response time window. Thus, a critical operational gap still persists between post-disaster mapping and the anticipatory planning required to strengthen the overall disaster resilience of the local community \citep{cai2018synthesis, ye2023developing}. Aligning FM's capabilities with localized preparedness initiatives will ensure that automated mapping not only guides immediate humanitarian recovery but also actively informs disaster risks and supports anticipatory disaster financing \citep{kull2016building}. 

To effectively close the widening ``Disaster Resilience Gap'', future works shall seek to go beyond post-disaster damage assessment to pre-disaster vulnerability perception (e.g., earthquakes, floods, and wildfires). Integrating disaster risk models with contemporary urban planning practice is essential to enable disaster preparedness and anticipatory resource planning, especially for the climate-vulnerable region and community \citep{cutter2008place}.

\section{Conclusions} \label{conclusion}

In this paper, we present Smart Transfer, a novel GeoAI framework leveraging pre-trained VFMs for rapid, cross-regional building damage mapping with post-disaster VHR imagery. By addressing annotation scarcity and cross-regional generalization issues, we demonstrate that the state-of-the-art VFMs, specifically DINOv3, substantially outperform traditional convolutional architectures (e.g., ResNet, YOLO) in building damage detection and classification, especially when transferring across disaster regions. 

As a core contribution, Smart Transfer designs two model transfer strategies: Pixel-wise Clustering (PC) and Distance-Penalized Triplet (DPT) to handle domain shift across diverse urban morphologies and damage features. Herein, PC ensures robust prototype-level global feature alignment w.r.t distinct urban morphology. DPT functions as a efficient spatial regularization of patch-level spatial autocorrelation patterns by assigning stronger penalties to semantically inconsistent yet spatially adjacent patches. Jointly trained with PC and DPT, Smart Transfer significantly improves the capability of vision FMs in learning more transferable structural knowledge with minimum training and limited source-domain samples. This feature is particularly important for real-world post-disaster response scenarios, providing a highly scalable, automated solution to accelerate situational awareness during the critical ``Golden 72 Hours''. 

Extensive experiments and ablations, using the recent 2023 Türkiye-Syria earthquake as a case study, confirms the competitive performance of Smart Transfer across multiple cross-region transfer settings, specifically LODO and SSDC, with minimum training and data annotation. More importantly, key findings in this work offers stimulating evidence towards an autonomous post-disaster damage mapping solution, which can be further integrating with anticipatory disaster preparedness and financing towards enhancing disaster resilience, especially for climate-vulnerable regions and communities.

\section*{Acknowledgments} 

This work was supported by the Start-Up Grant (SUG) project “Geospatial Artificial Intelligence for Climate Resilient Urban Environment” from the National University of Singapore (Grant No. E-109-00-0036-01), the Young Scientists Fund of the National Natural Science Foundation of China  (Grant No. 42401567), Guangdong Provincial Project (Grant No. 2024QN11G095), and the AI Research and Learning Base of Urban Culture, Guangdong Provincial Department of Education (Grant No. 2023WZJD008).

\section*{Disclosure Statement} No potential conflict of interest was reported by the authors.


\printcredits

\bibliographystyle{cas-model2-names}

\bibliography{interacttfvsample}



\end{document}